\documentclass{article}

\usepackage[numbers, sort, comma, square]{natbib}
\usepackage[preprint]{neurips_2025}

\usepackage[utf8]{inputenc} 
\usepackage[T1]{fontenc}    
\usepackage{hyperref}       
\usepackage{url}            
\usepackage{booktabs}       
\usepackage{amsfonts}       
\usepackage{nicefrac}       
\usepackage{microtype}      
\usepackage{xcolor}         

\usepackage{multirow}
\usepackage{tikz}
\usepackage{pgfplots}
\usepackage{pgfplotstable}
\usepackage{pgf-pie}
\usepackage{subcaption}
\usetikzlibrary{patterns, shapes, positioning}
\usepackage{algorithm}
\usepackage{algpseudocode}
\usepackage{amsmath}
\usepackage{xcolor}
\usepackage{colortbl}
\usepackage{enumitem}
\usepackage{wrapfig}
\usepackage{mdframed}
\usepackage{cleveref}
\usepackage{algorithm}
\usepackage{algpseudocode}
\usepackage{amsmath}
\usepackage{xcolor}
\usepackage{colortbl}
\usepackage{enumitem}
\usepackage{wrapfig}
\usepackage{tabularx}
\usepackage{listings}
\usepackage{makecell}
\usepackage{authblk}
\usepackage[most]{tcolorbox}

\newtcblisting{promptbox}[1][]{%
  enhanced,
  breakable,
  colback=gray!10,
  colframe=gray!50,
  arc=1mm,
  boxsep=4pt,
  fontupper=\ttfamily\small,
  title=#1,
  listing only,
  listing engine=listings,
  listing options={
    inputencoding=latin1,    
    breaklines=true,
    breakatwhitespace=false,
    basicstyle=\ttfamily\small,
    columns=fullflexible,
    escapeinside={(*@}{@*)}      
  }%
}
\title{Learning to Reason via Mixture-of-Thought for Logical Reasoning}

\author{%
  Tong Zheng\textsuperscript{1*}, 
  Lichang Chen\textsuperscript{1*}, 
  Simeng Han\textsuperscript{2}, 
  R. Thomas McCoy\textsuperscript{3}, 
  and Heng Huang\textsuperscript{1}\\
  \textsuperscript{*}These authors contributed equally.\\
  \textsuperscript{1}Dept.\ of Computer Science, UMD, College Park, MD 20742\\
  \textsuperscript{2}Dept.\ of Computer Science, Yale University, New Haven, CT 06520\\
  \textsuperscript{3}Dept.\ of Linguistics, Yale University, New Haven, CT 06520\\
}

\begin{document}

\maketitle

\begin{abstract}
Human beings naturally utilize multiple reasoning modalities to learn and solve logical problems, \textit{i.e.}, different representational formats such as natural language, code, and symbolic logic.
In contrast, most existing LLM-based approaches operate with a single reasoning modality during training, typically natural language.
Although some methods explored modality selection or augmentation at inference time, the training process remains modality-blind, limiting synergy among modalities.
To fill in this gap, we propose \textit{Mixture-of-Thought} (MoT), a framework that enables LLMs to reason across three complementary modalities: natural language, code, and a newly introduced symbolic modality, truth-table, which systematically enumerates logical cases and partially mitigates key failure modes in natural language reasoning. MoT adopts a two-phase design: (1) \textbf{self-evolving MoT training}, which jointly learns from filtered, self-generated rationales across modalities; and (2) \textbf{MoT inference}, which fully leverages the synergy of three modalities to produce better predictions. Experiments on logical reasoning benchmarks including FOLIO and ProofWriter demonstrate that our MoT framework consistently and significantly outperforms strong LLM baselines with single-modality chain-of-thought approaches,
achieving up to +11.7pp average accuracy gain.
Further analyses show that our MoT framework benefits both training and inference stages; that it is particularly effective on harder logical reasoning problems; and that different modalities contribute complementary strengths, with truth-table reasoning helping to overcome key bottlenecks in natural language inference.  The training codes are publicly
available on GitHub\footnote{\url{https://github.com/zhengkid/Truth_Table_Logical_Reasoning}}.
\end{abstract}

\section{Introduction}
Large language models (LLMs) have demonstrated remarkable progress in logical reasoning tasks, especially propelled by methods like Chain-of-Thought (CoT) prompting~\cite{wei2022chain}.
However, these CoT approaches predominantly rely on single reasoning modality, \textit{i.e.}, natural language, even when employing ensemble methods~\cite{li2023think,wang2025typedthinker,wang2022self,brown2024large,snell2025scaling,liang2023encouraging}.
Here we refer to a \textit{modality} as a distinct thought paradigm\footnote{We use the terms thought paradigm and reasoning modality interchangeably.} (\emph{e.g.} natural language, symbolic, or code), which differs in representation and inference process. 
On the other hand, neuro-symbolic methods~\cite{pan2023logic,olausson2023linc,ryu2025divide} utilize LLMs as translators and delegate reasoning to external symbolic solvers. Recent work combines CoT with symbolic reasoning via either selecting a single modality per instance~\cite{Xiong2024HYBRIDMINDMS} or augmenting one modality with the other—while keeping reasoning confined to symbolic~\cite{xu-etal-2024-faithful} or natural language~\cite{liu2024logic}. 
These methods combine modalities only during inference and ignore the synergy of different modalities during training, thus failing to fully exploit the complementary strengths of different reasoning modalities.

\definecolor{CodeColor}{HTML}{1f77b4}       
\definecolor{TruthTableColor}{HTML}{2ca02c} 
\definecolor{NLCoTColor}{HTML}{ff7f0e}      
\pgfplotsset{compat=1.18}
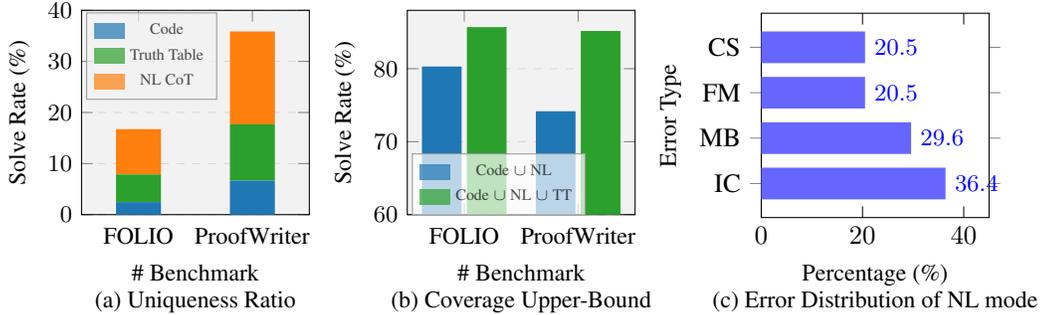
\begin{figure*}[t]
\centering
\vspace{-2em}
\begin{tikzpicture}
\node[anchor=south west] (a) at (0,0) {
\begin{tikzpicture}
\begin{axis}[
    width=0.33\textwidth,
    height=4.3cm,
    axis background/.style={fill=gray!10},
    xlabel={\small \makecell{\# Benchmark\\ (a) Uniqueness Ratio}},
    ylabel={\small Solve Rate (\%)},
    symbolic x coords={FOLIO, ProofWriter},
    xtick=data,
    ybar stacked,
    bar width=17pt,
    ymajorgrids=true,
    grid style={dashed,gray!30},
    enlarge x limits=0.5,
    tick align=inside,
    clip=false,
    ymin=0, ymax=40,
    tick label style={font=\small},
    label style={font=\small},
    x tick label style={align=center, yshift=-0.3ex},
    axis lines=box,
    legend style={
        at={(0.02,0.99)}, anchor=north west,
        font=\tiny,
        fill=white, fill opacity=0.7,
        draw=black!40,
        fill=gray!10,
        legend columns=1
    },
    legend image code/.code={
      \draw[draw=none, fill=#1] (0cm,-0.1cm) rectangle (0.4cm,0.1cm);
    }
]
\addplot+[fill=CodeColor, draw=none] coordinates {
    (FOLIO,2.46) (ProofWriter,6.67)};
\addplot+[fill=TruthTableColor, draw=none] coordinates {
    (FOLIO,5.41) (ProofWriter,11.0)};
\addplot+[fill=NLCoTColor, draw=none] coordinates {
    (FOLIO,8.87) (ProofWriter,18.2)};
\legend{Code, Truth Table, NL CoT}
\end{axis}
\end{tikzpicture}
};

\node[anchor=south west] (b) at (0.31\textwidth,0) {
\begin{tikzpicture}
\begin{axis}[
    width=0.33\textwidth,
    height=4.3cm,
    axis background/.style={fill=gray!10},
    xlabel={\small \makecell{\# Benchmark\\(b) Coverage Upper-Bound}},
    ylabel={\small Solve Rate\,(\%)},
    symbolic x coords={FOLIO,ProofWriter},
    xtick=data,
    ybar,
    bar width=15pt,
    ymajorgrids=true,
    grid style={dashed,gray!30},
    enlarge x limits=0.5,
    tick align=inside,
    clip=false,
    ymin=60, ymax=88,
    tick label style={font=\small},
    label style={font=\small},
    axis lines=box,
    x tick label style={align=center, yshift=-0.3ex},
    legend style={
      at={(0.02,0.30)}, anchor=north west,
      font=\tiny,
      fill=white, fill opacity=0.7,
      draw=black!40,
      legend columns=1
    },
    legend image code/.code={
      \draw[draw=none, fill=#1] (0cm,-0.1cm) rectangle (0.4cm,0.1cm);
    }
]
\addplot+[fill=CodeColor, draw=none] coordinates {
    (FOLIO,80.3) (ProofWriter,74.17)};
\addplot+[fill=TruthTableColor, draw=none] coordinates {
    (FOLIO,85.7) (ProofWriter,85.17)};
\legend{Code $\cup$ NL, Code $\cup$ NL $\cup$ TT}
\end{axis}
\end{tikzpicture}
};

\node[anchor=south west] () at (0.62\textwidth,0) {
\begin{tikzpicture}
  \begin{axis}[
    xbar,
    bar width=12pt,
    width=0.33\textwidth,
    height=4.3cm,
    xmin=0, xmax=45,
    xlabel={\small \makecell{Percentage (\%)\\(c) Error Distribution of NL mode }},
    symbolic y coords={
      IC,
      MB,
      FM,
      CS
    },
    ytick=data,
    ylabel={\small Error Type},
    nodes near coords,
    every node near coord/.append style={
      font=\small,
      xshift=0pt
    },
    axis lines=box,
    major x grid style={dashed,gray!30},
    major x tick style={black},
    enlarge y limits=0.25
  ]
    \addplot+[draw=none, fill=blue!60] coordinates {
      (36.4,IC)
      (29.6,MB)
      (20.5,FM)
      (20.5,CS)
    };
  \end{axis}
\end{tikzpicture}
};

\end{tikzpicture}

\vspace{-1em}
\caption{\footnotesize{(a) Qwen‑2.5‑7B‑Instruct solves $\simeq$20\% of FOLIO and $\simeq$35\% of ProofWriter exclusively per paradigm. (b) Code+NL+truth‑table yields higher upper‑bound coverage than code+NL alone~\cite{Xiong2024HYBRIDMINDMS}. (c) In NL modes, invalid‑converse (IC) and missing‑branch (MB) errors comprise $\simeq$66\% of failures (CS: commonsense injection; FM: factual misquote). Percentages sum to more than 100\% because some cases exhibit multiple error types. We provide illustrative examples in Appendix \ref{subsec:example_error_types}}}
\label{fig:motivation}
\vspace{-2em}
\end{figure*}
\par This limitation contrasts sharply with human cognition: Humans naturally employ multiple reasoning modalities, flexibly switching among natural language explanations, code-based procedural thinking, and formal symbolic manipulation, both when learning complex logical skills and when solving novel problems~\cite{newell1972human,gentner1983structure,larkin1987diagram, GOLDIN1998137}. This cognitive versatility, the ability to represent and process information in diverse formats, is crucial for robust reasoning and generalization. Current LLMs, largely confined to single-modality training and inference, lack this flexibility. It raises a critical question: \emph{Can LLMs achieve more robust and versatile logical reasoning by explicitly learning to operate across multiple complementary reasoning modalities?}

Addressing this question requires tackling two challenges:
1) It is still unclear which reasoning modalities should be included; the selected modalities must be complementary to make joint learning worthwhile.  
2) Equipping an LLM with multiple modalities is non-trivial, as large aligned reasoning trajectories are scarce. 
Our investigation reveals crucial insights for reasoning modality selection. 
\begingroup
\setlist[itemize]{leftmargin=10pt, itemsep=2pt, topsep=0pt, partopsep=0pt, parsep=0pt}
\begin{itemize}
    \item \textbf{Natural language bottleneck.} Figure \ref{fig:motivation} (c) shows that nearly two thirds of CoT errors arise from \emph{missing branches} and \emph{invalid converse}, \textit{i.e.}, poor exhaustive enumeration and complex inference (See examples in Appendix \ref{subsec:example_error_types}). Truth‑table reasoning, which systematically lists all possibilities, naturally complements this weakness; therefore, we incorporate a symbolic truth‑table paradigm.
    \item \textbf{Code–NL complementarity.}  Inspired by HybridMind~\cite{yue2024large,Xiong2024HYBRIDMINDMS}, where they show preliminary results that a code paradigm could complement NL reasoning, we also incorporate code as one reasoning modality into our framework.
    \item \textbf{Paradigm overlap \& uniqueness.} Figure \ref{fig:motivation} (a-b) shows that 35.8\% of ProofWriter items and 16.7\% of FOLIO items are solved by \emph{exactly one} paradigm, while the union of three reasoning modalities covers up to 85\% of all instances—evidence that combining NL, code, and truth‑table reasoning is necessary, outperforming the simple combination of code and natural language~\cite{Xiong2024HYBRIDMINDMS}. 
\end{itemize}
\endgroup
Building on these insights, we propose \textbf{M}ixture-\textbf{o}f-\textbf{T}hought (MoT), a human-inspired framework that enables LLMs to reason via three complementary reasoning modalities: natural language reasoning, code reasoning, and symbolic reasoning; an example is shown in \cref{example:MoT} to illustrate each modality.
It is worth noticing that we introduce a new truth-table-based symbolic reasoning that systematically grounds propositional variables, constructs a partial truth table by pruning assignments that violate any premise, and infers the final answer by checking whether the conclusion holds across the remaining assignments.
Our MoT consists of two parts.
One part is training: we propose a self-evolving MoT training algorithm, which improves the model's reasoning ability in each modality through joint iterative optimization (\Cref{fig:pipeline} (a)).
Another part is inference, where we generate responses under each modality and leverage a voting mechanism to produce the final answer (\Cref{fig:pipeline} (b)). This straightforward strategy allows the model to combine diverse perspectives and make more robust predictions, particularly in complicated instances.


Empirically, 
we show that across three base models—Gemma-2-2B-IT, Gemma-2-9B-IT, and Qwen-2.5-7B-Instruct—our MoT consistently surpasses the CoT baseline on ProofWriter~\cite{tafjord-etal-2021-proofwriter} and FOLIO~\cite{han-etal-2024-folio}, with an average accuracy gain of up to~\textbf{+11.7pp}.
Notably, our 9B‑parameter MoT matches the results of GPT‑4 + Logic‑LM on FOLIO.
Additional analyses show that 1) MoT training outperforms single‑thought training; 2) Mixture‑of‑Thought sampling yields a higher oracle upper bound than single-thought sampling under the same inference budget 3) The gains grow with problem difficulty: MoT helps most on depth‑5 and other harder problems; and 4) A fine‑grained error study reveals a key natural‑language bottleneck, \emph{i.e.}, missing branches and frequent invalid converse errors, while the truth‑table paradigm help resolve some cases of exactly these types.

\vspace{-12pt}
\section{The Mixture-of-Thought Framework}
\label{section:approach}
\vspace{-8pt}

In this section, we introduce 1) three complementary reasoning modalities for logical reasoning (Sec. \ref{subsec:thinking_paradigm}); 2) our self-evolving training framework that jointly improves these reasoning modalities (Sec. \ref{subsec:mot_training}); and 3) our mixture-of-thought inference strategy that combines diverse but complementary reasoning paths to derive robust final predictions (Sec. \ref{subsec:mot_inference}).

\begin{table*}[t]
\centering
\vspace{-1em}
\caption{\footnotesize{Illustration of the three complementary reasoning modalities, \emph{i.e.}, natural-language CoT, code-based reasoning, and truth-table reasoning. We provide the corresponding outputs of LLMs in \cref{sec:more_examples_model_outputs}. }}
\label{example:MoT}
\scriptsize
\renewcommand{\arraystretch}{1.0}
{
\begin{tabular}{@{}p{0.3\textwidth} p{0.32\textwidth} p{0.32\textwidth}@{}}
\toprule
\multicolumn{3}{@{}p{\textwidth}@{}}{\scriptsize{\textbf{Premise:} Peter Parker is either a superhero or a civilian. The Hulk wakes up when he is angry. If he wakes up, he will break a bridge. If a destroyer breaks a bridge, Peter is not a civilian. Peter wears a uniform when he is a superhero. Thor is a god. A god is not a destroyer. Thor will break a bridge when he is happy. If Thor is happy, the Hulk is angry.}} \\ 
\multicolumn{3}{@{}p{\textwidth}@{}}{\textbf{Question:} If Thor is happy, does Peter Parker wear a uniform?} \\
\midrule

\textbf{Natural Language Reasoning} & \textbf{Code-Based Reasoning (Abstract)} & \textbf{Truth Table Reasoning} \\
\midrule
\scriptsize
<nl\_cot>\newline
Step 1: given premises "If Thor is happy, the Hulk is angry." and "The Hulk wakes up when he is angry.", we can know "If Thor is happy, then hulk wakes up." \newline
... \newline
Step 5: given premise "Peter Parker is either a superhero or a civilian." and derivation "If Thor is happy, then Peter Parker is not a civilian", we can know "If Thor is happy, then Peter Parker is a superhero." \newline
Step 6: given premise "Peter Parker wears a uniform when he is a superhero." and derivation "If Thor is happy, then Peter Parker is a superhero.", we can know "If Thor is happy, then Peter Parker wears a uniform" \newline
<end\_of\_nl\_cot>\newline
<answer>\textbf{Answer: (A)}<end\_of\_answer>
&
\scriptsize
\vspace{-1.2em}
\begin{verbatim}
<code>
class Hulk:
    def __init__(self, angry): ...
    def wakes_up(self): ...
    def breaks_bridge(self): ...
class Thor:
    def __init__(self, happy): ...
    def breaks_bridge(self): ...
class PeterParker:
    def __init__(self, is_superhero): 
        ...
    def wears_uniform(self): ...
def apply_premises(thor, hulk, peter): 
    ...
def run_inference(thor, hulk, peter): 
    ...
def check_conclusion(...): ...
thor = Thor(happy=True)
hulk = Hulk(angry=False)
peter = PeterParker(...)
result = check_conclusion(...)
\end{verbatim}
\vspace{-0.8em}
<end\_of\_code>\newline
<answer>\textbf{Answer: (A)}<end\_of\_answer>
&
\scriptsize
<truth\_table>\newline
\textbf{Let:} $T$ = Thor happy, $H$ = Hulk angry, $A$ = wakes up, $B$ = bridge breaks, $C$ = Peter is civilian, $S$ = superhero, $U$ = wears uniform. \newline

\textbf{Premises:} \newline
$S \lor C$, $H \rightarrow A$, $A \rightarrow B$, $T \rightarrow H$, $T \rightarrow B$, $B \rightarrow \lnot C$, $S \rightarrow U$. \newline

\textbf{Logical Chain (assume $T=\text{True}$):} \newline
$T \Rightarrow H \Rightarrow A \Rightarrow B$ \newline
$B \Rightarrow \lnot C \Rightarrow C = \text{False}$ \newline
$S \lor C \Rightarrow S = \text{True} \Rightarrow U = \text{True}$ \newline

\textbf{Truth Table:} \newline

\setlength{\tabcolsep}{1pt}
\resizebox{0.8\linewidth}{!}{
\begin{tabular}{ccccccc}
\toprule
$T$ & $H$ & $A$ & $B$ & $C$ & $S$ & $U$ \\
\midrule
True & True & True & True & False & True & True \\
\bottomrule
\end{tabular}
} \newline
\newline
<end\_of\_truth\_table>\newline
<answer>\textbf{Answer: (A)}<end\_of\_answer>
\\
\bottomrule
\end{tabular}}
\vspace{-0.5em}
\end{table*}

\vspace{-5pt}
\subsection{Human-Inspired Complementary Reasoning Modalities}
\label{subsec:thinking_paradigm}
\vspace{-5pt}
Drawing inspiration from human cognition and error analysis in Figure \ref{fig:motivation}, we argue that no single reasoning modality suffices for all logical challenges. Therefore, we equip a single model with three complementary modalities: natural language CoT, code CoT, and truth table CoT. Specifically, because natural‑language CoT often misses branches or makes invalid‑converse errors, we design a truth‑table approach that explicitly enumerates truth assignments and thus complements these weaknesses. Table \ref{example:MoT} illustrates how the three modalities solve a representative problem.
\begingroup
\setlist[itemize]{leftmargin=10pt, itemsep=2pt, topsep=0pt, partopsep=0pt, parsep=0pt}
\begin{itemize} 
    \item \textbf{Natural Language CoT:} The model explains its reasoning in plain natural language, decomposing the problem into step-by-step justifications. This format is flexible and interpretable.
    \item \textbf{Code CoT:} The model first transforms a logical problem to a PYTHON code and then derives the answer based on the PYTHON code. 
    We do not execute the code; instead, 
    we treat it as a way to describe logic in a structured form.
    \item \textbf{Truth Table CoT:} The model first explicitly generates a truth table by defining predicates based on the premises and conclusion, then enumerating possible truth assignments, and finally checking which ones satisfy the conclusion.
\end{itemize}
\endgroup
These complementary modalities are jointly exploited in our self-evolving training (Sec. \ref{subsec:mot_training}) and majority‑vote inference (Sec. \ref{subsec:mot_inference}). We now detail the design of the Truth Table CoT approach.

\paragraph{Truth‑Table CoT: Challenges and Design.} Two main challenges arise when enabling LLMs to reason with truth tables: 1) Exponential blow‑up: the number of rows grows exponentially with the propositional variables, easily exceeding the context window and compute budget; 2) First‑order grounding: practical tasks are given in first‑order logic; one must ground variables, select a finite predicate set, and still ensure that the resulting (partial) truth table remains tractable. To address these challenges, we propose a two‑step strategy: (i) grounding, which instantiates first‑order formulas into a finite set of propositional predicates~\cite{clarke2001bounded,wittocx2010grounding}, and (ii) reason to prune, which eliminates rows that violate any premise through reasoning via LLMs, keeping partial truth table (see Table \ref{example:MoT} and Appendix \ref{subsec:example_tt}). Finally, the LLMs derive the final output with the following rule: True if every surviving assignment satisfies the conclusion, False if none do, and Uncertain otherwise. Moreover, we assign modality-specific tags (e.g., \texttt{<code>} ... \texttt{<end\_of\_code>}) to explicitly indicate the format during training and inference. The prompts are detailed in Appendix~\ref{app:full_prompts}.

\begin{figure}[t!]
    \centering
    \vspace{-2em}
    \includegraphics[width=0.85\linewidth]{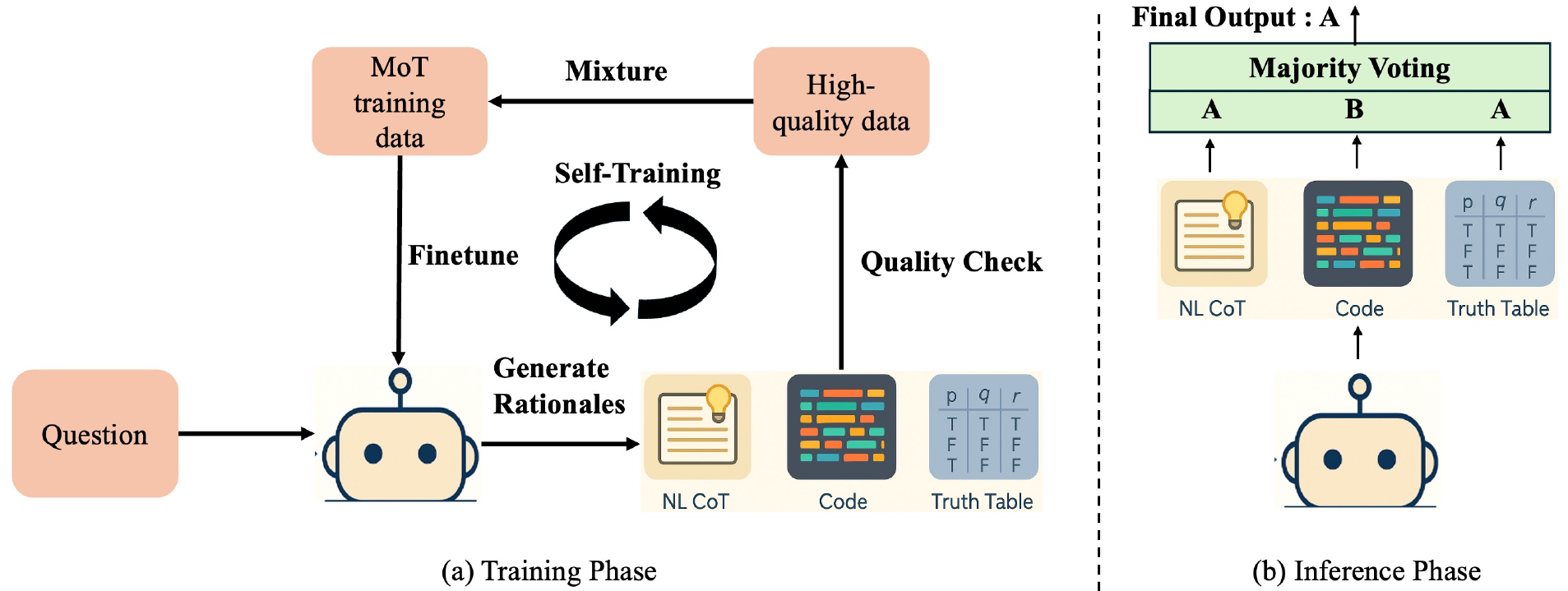}
    \caption{\footnotesize{Illustration of our MoT Framework. (a) \textbf{Training phase} with three key steps: 1) \textbf{Rationale Generation} where given an initial seed dataset, LLM generates rationales across reasoning modalities (NL, Code, and Truth Table); 2) \textbf{Quality Checking and Merging} where generated rationales are checked for correctness and format consistency, then merged into high-quality MoT training data; and 3) \textbf{Finetuning} where the model is trained using the MoT data. These steps iteratively repeats, forming a self-evolving training cycle. (b) \textbf{Inference phase:} the trained model generates outputs for each reasoning modality and applies majority voting to yield the final prediction (e.g., A).
}}
    \label{fig:pipeline}
\vspace{-1em}
\end{figure}

\subsection{Self-Evolving Mixture-of-Thought Training}
\label{subsec:mot_training}
Explicitly learning to reason across multiple complementary modalities, such as natural language, code, and symbolic truth tables, is non-trivial. A key challenge lies in the lack of annotated reasoning trajectories for each modality, especially for our newly introduced truth-table approach.
Collecting labeled CoT traces for all of these modalities is also costly and often infeasible.
To address this, we propose a self-evolving MoT training approach, which enables the model to operate across multiple complementary reasoning modalities by iteratively learning from its own generated reasoning traces. 

Given the policy $M$, our goal is to maximize the following objective across the problems $x$
and modalities $\mathcal{T} \in \{\mathrm{NL}, \mathrm{Code}, \mathrm{TruthTable}\}$ :
\begin{equation}
\sum_{i, t}\mathbb{E}_{\,(x_i, y_i)\sim \mathcal{D},\,t \sim \mathcal{T},\,(z_i^t, \hat{y}_i^t)\sim M(\cdot|x_i, t, \mathcal{E}_t)}
       \bigl[R(z_i^t,\hat{y}_i^t, y_i;t)\bigr],
\label{eq: objective}
\end{equation}
where $\mathcal{D} = \{(x_i, y_i)\}_{i=1}^D$ is the dataset with problem $x_i$ and corresponding ground-truth $y_i$;
$z^t_i$ and $\hat{y}^t_i$ be model-generated reasoning trace/answer with modality $t$ for $i$-th problem.
To elicit the reasoning modality $t$, we design a small few-shot example set $\mathcal{E}_t$ for each $t$, and prepend the exemplar from the set to each problem $x_i$.
Conditioned on $(x_i, t, \mathcal{E}_t)$, $z_i$ is sampled from policy $M$, followed by the prediction of the final answer $\hat{y}_i$. 
$R$ is the reward function and the design is detailed in the following.


\paragraph{Reward Function $R$.} In preliminary experiments, we observe mismatch between tags and reasoning traces. This error leads to performance degradation, as different modalities negatively interfere with each other. Notably, this error is especially prevalent in the code modality. 
We define the reward as:
\begingroup
\begin{align}
R(z, \hat{y}, y; t)
=\begin{cases}
1, & y=\hat{y} \;\land\; \mathrm{isValid}(z,t),\\
0, & \text{otherwise},
\end{cases}
\end{align}
\endgroup
where the isValid function checks the format consistency by two standards: a) each trace should correctly include its modality’s structural tag (e.g., \texttt{<end\_of\_nl\_cot>} for nl) and b) for code traces, ensuring the presence of both a valid function definition (\texttt{def}) and a class definition (\texttt{class});
Following \citep{zelikman2022star}, we also use the final answer to filter out the traces without the correct answer.

\paragraph{Training.} We conduct multiple rounds of self-evolving training until performance saturates. 
$M_n$ is used to denote the policy in the $n$-th round with trainable parameters $\theta_n$.
Leveraging the policy‑gradient~\cite{NIPS1999_464d828b} trick, we can easily obtain the gradient of \cref{eq: objective} as
\begin{equation}
\nabla J = \sum_{i, t}\mathbb{E}_{\substack{
(x_i, y_i) \sim \mathcal{D},
t \sim \mathcal{T} \\
(z_i^t, \hat{y}_i^t) \sim M_{n-1}(\cdot \mid x_i, t, \mathcal{E}_t)
}}
\left[
R(z_i^t,\hat{y}_i^t, y_i;t)\,
\nabla_{\theta_{n-1}} \log M_{n-1}(z_i^t, \hat{y}_i^t \mid x_i, t, \mathcal{E}_t)
\right].
\end{equation}

\begin{wrapfigure}[26]{r}{0.55\textwidth}
  \centering
  \vspace{-2em}
  \begin{minipage}{\linewidth}
\begin{algorithm}[H]
\caption{Self-Evolving MoT Training}
\label{alg:mot}
\textbf{Input:} an LLM $M$; dataset $\mathcal{D} = \{(x_i, y_i)\}_{i=1}^{D}$; reasoning modality $\mathcal{T} = \{\text{NL}, \text{Code}, \text{TruthTable}\}$, Sampling times $S$, few-shot examples $\mathcal{E} = \{\mathcal{E}_\mathrm{NL}, \mathcal{E}_\mathrm{Code}, \mathcal{E}_\mathrm{Truth Table} \} $  \\
\textbf{Output:} Mixture-of-Thought enhanced model $M_N$
\begin{algorithmic}[1]
\State $M_0 \gets M$ 
\For{$n = 1$ to $N$} 
    \State Initialize $\mathcal{D}_\mathrm{all,n}^{\text{gen}} \gets \emptyset$; $\mathcal{D}_\mathrm{NL,n}^{\text{gen}} \gets \emptyset$; $\mathcal{D}_\mathrm{Code,n}^{\text{gen}} \gets \emptyset$; $\mathcal{D}_\mathrm{TruthTable,n}^{\text{gen}} \gets \emptyset$
    \ForAll{$t \in \mathcal{T}$} 
        \For{$i = 1$ to $D$}
            \If{n = 1} 
            \State ${z}_i^t, \hat{y}_i^t \gets M_{n-1}(x_i; {t}; \mathcal{E}_t; S)$ 
            \Else 
            \State ${z}_i^t, \hat{y}_i^t \gets M_{n-1}(x_i; t; S)$ 
             \EndIf
            \If{$R({z}_i^t, \hat{y}_i^t, y_i; t)$=1}
                \State $\mathcal{D}_{t,n}^{\text{gen}} \gets \mathcal{D}_{t,n}^{\text{gen}} \cup \{(x_i, {z}_i^t, y_i)\}$
            \EndIf
        \EndFor
    \EndFor
    \State $\mathcal{D}_\mathrm{all,n}^{\text{gen}} \gets \mathrm{Mix}(\mathcal{D}_\mathrm{NL,n}^{\text{gen}}, \mathcal{D}_\mathrm{Code,n}^{\text{gen}}, \mathcal{D}_\mathrm{TruthTable,n}^{\text{gen}})$ 
    \State $M_n \gets \text{Train}(M_{n-1}, \mathcal{D}_\mathrm{all,n}^{\text{gen}})$
\EndFor
\State \Return $M_N$
\end{algorithmic}
\end{algorithm}
\end{minipage}
\end{wrapfigure}
Algorithm~\ref{alg:mot} and Figure~\ref{fig:pipeline} illustrate our multi-round training procedure.
At round $n$, we prompt the model $M_{n-1}$ to generate a reasoning trace ${z}_i^t$ and a predicted answer $\hat{y}_i^t$ for each $x_i$ across all reasoning modalities $t \in \mathcal{T}$ (Line 4-9).
It is worth noting that we use few‑shot prompting only in the first round (Line 7); once the model has bootstrapped its own reasoning ability, all subsequent rounds run in zero‑shot mode without additional exemplars (Line 9). 
We retain a sample only if it passes the quality filter (Line 11-13) and merge all surviving traces into $\mathcal D^{\text{gen}}_{\text{all},n}$ (Line 16).
The updated model $M_n$, which is finetuned from $M_{n-1}$ on the filtered dataset $\mathcal{D}^{\text{gen}}_{\text{all},n}$ (Line 17).
Unlike \cite{zelikman2022star}, which restarts from the base model each round, our training proceeds on-policy—learning from its own validated outputs. We demonstrate the effectiveness of this change in Appendix \ref{subsec:ablation_studies}.

\subsection{Mixture-of-Thought Inference}
\label{subsec:mot_inference}
To leverage the complementary strengths of three modalities, for each problem, we have three outputs corresponding to three modalities elicited by tagging, \textit{i.e.}, \texttt{<nl\_cot>}, \texttt{<code>}, and \texttt{<truth\_table>},  then we apply majority voting over outputs to decide the final answer.
In case of ties, we randomly pick up the answer from one reasoning modality.
We further explore the test-time scaling of MoT, and analyze its effectiveness in \cref{sec:test_time_scaling}.

\section{Empirical Evaluations}
\label{sec:experiments}
\subsection{Experimental Setups}
\paragraph{Models.} 
To validate the effectiveness of our MoT, 
we select three widely-used LLMs across different sizes and model families: Qwen-2.5-7B-Instruct~\cite{yang2024qwen2} and Gemma-2-2B-It/Gemma-2-9B-It~\cite{team2024gemma} as base models. These models span different sizes and capacities, ensuring diverse evaluation.
\paragraph{Baselines.} Our approach is a kind of chain-of-thought approach. To this end, we select baselines from two folds: 1) neuro-symbolic approach and 2) chain-of-though approach. In the first category, we select Logic-LM~\cite{pan2023logic} as a comparison. For the CoT approach, we select CoT~\cite{wei2022chain} as a comparison. Since these approaches heavily rely on strong instruction-following capabilities, we directly cite their performance results from the original papers based on GPT-4.  

\paragraph{Dataset.} We select two logical reasoning benchmarks: ProofWriter~\cite{tafjord-etal-2021-proofwriter} and FOLIO~\cite{han-etal-2024-folio} for evaluation. ProofWriter is known for its numerous test cases. We select the hardest subset, which consists of 600 questions with reasoning depths of 5, the same as ~\citet{pan2023logic}. FOLIO is recognized for its high-quality export-made realistic test cases with more diverse reasoning depths ranging from 1-8. It consists of 203 questions. We utilize accuracy and pass@k as metrics.

\begin{table}[t!]
\vspace{-1em}
\centering
\caption{\footnotesize{Accuracy (\%) on the FOLIO and ProofWriter benchmarks. Our MoT training consistently improves the performance of each base model. Applying MoT inference further enhances performance across both benchmarks and all models. @3 denotes Self-Consistency approach~\cite{wang2022self} with three votes. We provide full results of baselines across reasoning modalities in Appendix \ref{subsec:detailed_baseline_MoT_results_across_modalities} \& \ref{subsec:more_baseline_results}.}}
\vskip 0.1 in
\label{tab:comparison_final}
\resizebox{\textwidth}{!}{\begin{tabular}{l l l c c c }
\toprule
\textbf{Model} & \textbf{Method Type} & \textbf{Reasoning Modality} & 
\textbf{FOLIO} & \textbf{ProofWriter} & \textbf{Avg}  \\
\midrule
\multicolumn{6}{c}{\textbf{(A) Prior SOTA Approach}} \\
\multirow{2}{*}{GPT-4} 
 & \textit{Logic-LM} & - & 78.9 & 79.7 & 79.3 \\
 & \textit{CoT (Vanilla)} & - &  70.6  & 68.1 & 69.4 \\
\midrule
\multicolumn{6}{c}{\textit{\textbf{(B) Base Model: Gemma-2-2B-It}}} \\
Gemma-2-2B-It (3-Shot) & Single-Thought & Best (nl)        & 42.4 & 39.8  & 41.1 \\
Gemma-2-2B-It @ 3 (3-Shot) & Single-Thought & Best (nl)        & 45.3 & 38.8  & 42.1 \\
\textbf{MoT} (0-Shot) & Single-Thought & Best         & 61.1 & 62.7  & 61.9 \\
\textbf{MoT} (0-Shot) & Mixture-of-Thought & All         & \bf 62.6 & \bf 65.0  & \bf 63.8 \\
\midrule
\multicolumn{6}{c}{\textit{\textbf{(C) Base Model: Gemma-2-9B-It}}} \\
Gemma-2-9B-It (3-shot) & Single-Thought & Best (nl)        & 69.5  & 61.2 & 65.4 \\
Gemma-2-9B-It @ 3 (3-shot) & Single-Thought & Best (nl)         & 72.9  & 62.7 & 67.8 \\
\textbf{MoT} (0-shot) & Single-Thought & Best         & 76.9  & 69.5 & 73.2 \\
\textbf{MoT} (0-shot) & Mixture-of-Thought & All         & \bf 78.9 & \bf 70.7  & \bf 74.8 \\
\midrule
\multicolumn{6}{c}{\textit{\textbf{(D) Base Model: Qwen2.5-7B-Instruct}}} \\
Qwen2.5-7B-Instruct (3-shot) & Single-Thought & Best (nl)         & 71.9 & 60.5  & 66.2 \\
Qwen2.5-7B-Instruct @ 3 (3-shot) & Single-Thought & Best (nl)         & 73.4  & 65.8 & 69.6  \\
\textbf{MoT} (0-shot) & Single-Thought & Best         & 75.9  & 69.2   & 72.6 \\
\textbf{MoT}  (0-shot) & Mixture-of-Thought & All         & \textbf{ 78.3 } & \textbf{ 71.8} & \bf 75.1  \\
\bottomrule
\end{tabular}}
\end{table}

\paragraph{Training/Inference Details.} For each dataset, we collect 1000 training samples from the training set. We perform 2 or 3 rounds of self-evolving training. In each round, the model is fine-tuned for two epochs using a learning rate of 2e-5 and a batch size of 128. During the trajectory collection phase, the temperature, max\_tokens, and sample count are set to 1.0, 2048, and 10, respectively. We sample each problem $10$ times during trajectory collection to maximize coverage. Of all the generated traces, only the first single trajectory that satisfies our quality criteria is retained for the final training set. For evaluation, the temperature and max\_tokens are configured to 0.7 and 2048, respectively. We run all experiments on 4 H100 GPUs. We employ vLLM engine~\cite{kwon2023efficient} to improve inference efficiency.
\subsection{Main Results}
\label{sec:main_results}

\Cref{tab:comparison_final} displays the results on FOLIO and ProofWriter benchmarks. First, our Mixture‑of‑Thought (MoT) training with Single‑Thought inference outperforms the corresponding base models by an average of \textbf{11.7pp} (from 41.1\% to 61.9\% for Gemma-2-2b-It, from 65.4\% to 73.2\% for Gemma-2-9b-It and from 66.2\% to 72.6\% for Qwen-2.5-7b-Instruct), demonstrating the effectiveness of our training strategy.
When we further apply MoT inference, the MoT‑trained model yields consistent additional gains of up to \textbf{2.0pp}.
Notably, our 9B model achieves 78.9\% accuracy on FOLIO, matching the performance of Logic-LM, which uses an external solver and focuses on close-sourced SoTA LLMs.
We provide a detailed performance of both base models and the corresponding MoT models, as well as stronger baselines, in Appendix \ref{subsec:detailed_baseline_MoT_results_across_modalities} \& \ref{subsec:more_baseline_results}.


\subsection{Mixture-of-Thought Training vs. Single-Thought Training}
\label{sec:MoT_Training_Exp}
In this section, we try to answer the key question: \textit{Does MoT training truly offer benefits over Single‑Thought training?}  We have two baselines: 1) models trained on single-thought data and 2) models trained on partially MoT data, e.g., Code + NL.
We evaluate both in‑mode accuracy and cross‑mode generalization.
To enhance model's format following ability, we use 3‑shot prompting to make model output the specific reasoning modality.
Table \ref{tab:folio_accuracy_sot_vs_mot} illustrates the results on FOLIO benchmark.

\begin{table}[t!]
\vspace{-2em}
\setlength{\tabcolsep}{3pt}
\caption{\footnotesize{Accuracy (\%) of different training strategies across reasoning modalities (Same Round). 
Shaded cells denote in-domain evaluation, i.e., testing on the same modalities during training. Avg. refers to the average performance using three modalities while Ensemble means the majority vote results on three modalities.
Values underlined indicate that the model did not follow the instruction (\emph{e.g.}, when asked to use Code, it still used NL).}}
\label{tab:folio_accuracy_sot_vs_mot}
\centering
\vskip 0.1in
\resizebox{\textwidth}{!}{\begin{tabular}{lccccccc}
\toprule
\textbf{Training Approach } & \textbf{Param} & \textbf{Data} & \textbf{Code} & \textbf{NL\_CoT} & \textbf{Truth Table} & \bf Avg. & \textbf{Ensemble}\\
\midrule
\textit{w/o Training} \\
\midrule
\quad - & 9B & N/A & 56.7 & 69.5  & 63.6 & 63.3 & 66.0 \\
\midrule
\textit{Single-Thought Training} \\
\midrule
\quad  Single-Thought (Code)        & 9B & -    & \cellcolor{green!35}61.6 & 59.1  & 64.0 & 61.6  & 70.4\\
\quad  Single-Thought (NL\_CoT)    & 9B & -    & 52.7          &  \cellcolor{blue!35}73.9 & 69.0 & 65.2 & 73.4\\
\quad  Single-Thought (Truth Table) & 9B  & - & 53.2         & 69.0  & \cellcolor{orange!35}\bf 71.9 & 64.7 & 71.9 \\
\quad  Single-Thought (Three Models Combined)    & 3x9B & {$\sim 3\times$}& \cellcolor{red!35}61.6         & \cellcolor{red!35}73.9  & \cellcolor{red!35}\bf 71.9 & 69.1 & 77.3 \\
\midrule
\textit{Mixture-of-Thought Training} \\ 

\quad  Mixture-of-Thought (NL\_CoT + Truth Table)  & 9B  & {$\sim 2\times$}  &    \underline{65.5}       & \cellcolor{red!35} 72.9  & \cellcolor{red!35}69.5 & 69.3 & 72.9\\
\quad  Mixture-of-Thought (Truth Table + Code)    & 9B &  {$\sim 2\times$} & \cellcolor{red!35}70.0         & 71.4 & \cellcolor{red!35} 62.1  & 67.8 & 72.4 \\
\quad  Mixture-of-Thought (Code + NL\_CoT)   & 9B &  {$\sim 2\times$}  & \cellcolor{red!35} 70.9         & \cellcolor{red!35}70.0  & \underline{74.4} & 71.8 & 74.9 \\
\quad Mixture-of-Thought (Default, All)   & 9B &  {$\sim 3\times$}  & \cellcolor{red!35} \bf73.9          & \cellcolor{red!35}\bf 76.9  & \cellcolor{red!35}70.0 & \bf 73.6 & \bf 78.9\\
\bottomrule
\end{tabular}}
\vspace{-1em}
\end{table}

\paragraph{SoT vs. MoT.} 
First, MoT training achieves the highest average accuracy across all three modalities, beating single-thought trained model, which indicates that our MoT training can jointly improve reasoning ability across all modalities.
Second, MoT training can further push the performance boundary for each reasoning modality.
For example, by using two of the three modalities, \emph{i.e.}, Code and NL\_CoT, the trained models outperform all single-thought baselines.
This clearly indicates synergy between these three complementary modalities during training.
Third, deploying one model for each modality is resource-expensive.
In contrast, MoT training enables a single model to seamlessly switch among reasoning modalities based on prompts.

\paragraph{Partial MoT vs. MoT} Our default Mixture-of-Thought setting yields the best average performance and achieves the best accuracy by using two combined reasoning paradigms, which indicates that all the modalities are useful.
This superiority is further reflected in the ensemble accuracy, where MoT achieves 78.9\%.  We provide more evidence in Sec. \ref{sec:further_analysis_error_overlap} and Appendix \ref{app:complementary_uniqueness_analysis}.

\paragraph{Additional Ablations for MoT Training} We further give more analysis to show 1) robust and optimal design of the MoT framework (Appendix \ref{subsec:ablation_studies}); 2) MoT training is better than single-thought training with distillation data  (Appendix \ref{subsec:effect_data_quality}) and 3) MoT data outperform an equivalent amount of diverse single-thought CoT data (Appendix \ref{subsec:diverse_SoT_vs_MoT}). These results underscore the practical and broader value of our MoT framework.

\subsection{Test-Time Scaling Across Reasoning Modalities}
\label{sec:test_time_scaling}
\definecolor{darkgreen}{RGB}{0,100,0}
\definecolor{tableauBlue}{RGB}{31,119,180}
\definecolor{tableauOrange}{RGB}{255,127,14}
\definecolor{tableauGreen}{RGB}{44,160,44}
\definecolor{tableauRed}{RGB}{214,39,40}
\definecolor{tableauPurple}{RGB}{148,103,189}
\pgfplotsset{compat=1.18}
\usepgfplotslibrary{groupplots,fillbetween}
\pgfplotsset{every axis/.append style={font=\small}}

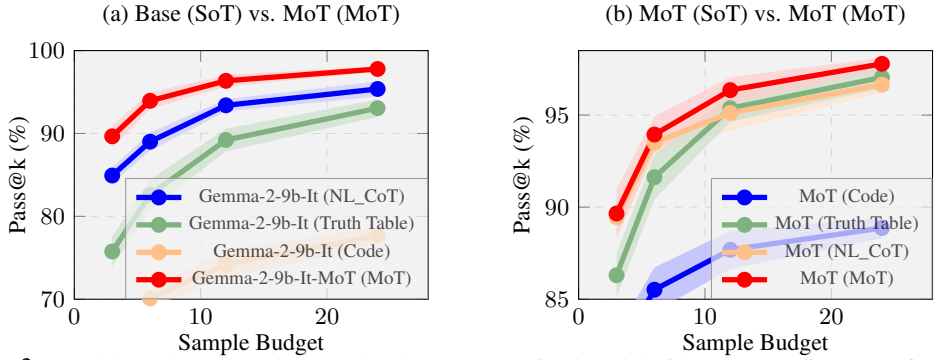
\begin{figure*}[t!]
  \centering
  \begin{tikzpicture}[]
    \begin{groupplot}[
      group style={group size=2 by 1, horizontal sep=2cm},
      axis background/.style={fill=gray!10},
      width=0.45\textwidth,
      height=0.35\textwidth,
      xlabel={Sample Budget},
      ylabel={Pass@k (\%)},
      xmin=0, xmax=28,
      ymin=70, ymax=100,
      grid=both,
      grid style={dashed,gray!30},
      xlabel style={at={(axis description cs:0.5,-0.1)}, anchor=north},
      ylabel style={at={(axis description cs:-0.1,0.5)}, anchor=south}
    ]
      \nextgroupplot[
        title={(a) Base (SoT) vs. MoT (MoT) },
        legend style={at={(axis description cs:1,0)}, anchor=south east, font=\small, draw=black!40, fill=gray!10, fill opacity=0.7}
      ]
      
      \addplot+[name path=base, color=blue, line width=2pt, mark=*, mark size=2pt, mark options={fill=none}]
        table [x index=0, y index=1] {data/FOLIO_Baseline_NL.dat};
      \addlegendentry{\scalebox{0.8}{Gemma-2-9b-It (NL\_CoT)}}

      \addplot+[name path=base_tt, color=darkgreen!50, line width=2pt, mark=*, mark size=2pt, mark options={fill=none}]
        table [x index=0, y index=1] {data/FOLIO_Baseline_TT.dat};
      \addlegendentry{\scalebox{0.8}{Gemma-2-9b-It (Truth Table)}}

      \addplot+[name path=base_code, color=tableauOrange!50, line width=2pt, mark=*, mark size=2pt, mark options={fill=none}]
        table [x index=0, y index=1] {data/FOLIO_Baseline_Code.dat};
      \addlegendentry{\scalebox{0.8}{Gemma-2-9b-It (Code)}}


      \addplot+[name path=moth, color=red, line width=2pt, mark=*, mark size=2pt, mark options={fill=none}]
        table [x index=0, y index=1] {data/FOLIO_MoT.dat};
      \addlegendentry{\scalebox{0.8}{Gemma-2-9b-It-MoT (MoT)}}

      \addplot+[name path=baseupper, draw=none, mark=none]
        table [x index=0, y expr=\thisrowno{1}+\thisrowno{2}] {data/FOLIO_Baseline_NL.dat};
      \addplot+[name path=baselower, draw=none, mark=none]
        table [x index=0, y expr=\thisrowno{1}-\thisrowno{2}] {data/FOLIO_Baseline_NL.dat};
      \addplot [blue!20, opacity=0.7, forget plot, on layer=axis background]
        fill between[of=baseupper and baselower];

    \addplot+[name path=basettupper, draw=none, mark=none]
        table [x index=0, y expr=\thisrowno{1}+\thisrowno{2}] {data/FOLIO_Baseline_TT.dat};
      \addplot+[name path=basettlower, draw=none, mark=none]
        table [x index=0, y expr=\thisrowno{1}-\thisrowno{2}] {data/FOLIO_Baseline_TT.dat};
      \addplot [darkgreen!20, opacity=0.7, forget plot, on layer=axis background]
        fill between[of=basettupper and basettlower];

    \addplot+[name path=basecodeupper, draw=none, mark=none]
        table [x index=0, y expr=\thisrowno{1}+\thisrowno{2}] {data/FOLIO_Baseline_Code.dat};
      \addplot+[name path=basecodelower, draw=none, mark=none]
        table [x index=0, y expr=\thisrowno{1}-\thisrowno{2}] {data/FOLIO_Baseline_Code.dat};
      \addplot [tableauOrange!30, opacity=0.7, forget plot, on layer=axis background]
        fill between[of=basecodeupper and basecodelower];

      
      \addplot+[name path=mothupper, draw=none, mark=none]
        table [x index=0, y expr=\thisrowno{1}+\thisrowno{2}] {data/FOLIO_MoT.dat};
      \addplot+[name path=mothlower, draw=none, mark=none]
        table [x index=0, y expr=\thisrowno{1}-\thisrowno{2}] {data/FOLIO_MoT.dat};
      \addplot [red!20, opacity=0.7, forget plot, on layer=axis background]
        fill between[of=mothupper and mothlower];
      
      \nextgroupplot[
        title={(b) MoT (SoT) vs. MoT (MoT)},
        legend style={at={(axis description cs:1,0)}, anchor=south east, font=\small, draw=black!40, fill=gray!10, fill opacity=0.7},
        xmin=0, xmax=28,
      ymin=85, ymax=98.5,
      ]

      \addplot+[name path=moth_code_nl, color=blue, line width=2pt, mark=*, mark size=2pt, mark options={fill=none}]
        table [x index=0, y index=1] {data/FOLIO_Code.dat};
      \addlegendentry{\scalebox{0.8}{MoT (Code)}}


      \addplot+[name path=moth_code_tt, color=darkgreen!50, line width=2pt, mark=*, mark size=2pt, mark options={fill=none}]
        table [x index=0, y index=1] {data/FOLIO_TT.dat};
      \addlegendentry{\scalebox{0.8}{MoT (Truth Table)}}


      \addplot+[name path=moth_nl_tt, color=tableauOrange!50, line width=2pt, mark=*, mark size=2pt, mark options={fill=none}]
        table [x index=0, y index=1] {data/FOLIO_NL.dat};
      \addlegendentry{\scalebox{0.8}{MoT (NL\_CoT)}}

      \addplot+[name path=moth, color=red, line width=2pt, mark=*, mark size=2pt, mark options={fill=none}]
        table [x index=0, y index=1] {data/FOLIO_MoT.dat};
      \addlegendentry{\scalebox{0.8}{MoT (MoT)}}

      \addplot+[name path=baseupper, draw=none, mark=none]
        table [x index=0, y expr=\thisrowno{1}+\thisrowno{2}] {data/FOLIO_NL.dat};
      \addplot+[name path=baselower, draw=none, mark=none]
        table [x index=0, y expr=\thisrowno{1}-\thisrowno{2}] {data/FOLIO_NL.dat};
      \addplot [tableauOrange!20, opacity=0.7, forget plot, on layer=axis background]
        fill between[of=baseupper and baselower];

    \addplot+[name path=basettupper, draw=none, mark=none]
        table [x index=0, y expr=\thisrowno{1}+\thisrowno{2}] {data/FOLIO_TT.dat};
      \addplot+[name path=basettlower, draw=none, mark=none]
        table [x index=0, y expr=\thisrowno{1}-\thisrowno{2}] {data/FOLIO_TT.dat};
      \addplot [darkgreen!20, opacity=0.7, forget plot, on layer=axis background]
        fill between[of=basettupper and basettlower];

    \addplot+[name path=basecodeupper, draw=none, mark=none]
        table [x index=0, y expr=\thisrowno{1}+\thisrowno{2}] {data/FOLIO_Code.dat};
      \addplot+[name path=basecodelower, draw=none, mark=none]
        table [x index=0, y expr=\thisrowno{1}-\thisrowno{2}] {data/FOLIO_Code.dat};
      \addplot [blue!30, opacity=0.7, forget plot, on layer=axis background]
        fill between[of=basecodeupper and basecodelower];

      
      \addplot+[name path=mothupper, draw=none, mark=none]
        table [x index=0, y expr=\thisrowno{1}+\thisrowno{2}] {data/FOLIO_MoT.dat};
      \addplot+[name path=mothlower, draw=none, mark=none]
        table [x index=0, y expr=\thisrowno{1}-\thisrowno{2}] {data/FOLIO_MoT.dat};
      \addplot [red!20, opacity=0.7, forget plot, on layer=axis background]
        fill between[of=mothupper and mothlower];

    \end{groupplot}
  \end{tikzpicture}
 \vspace{-1em}
 \caption{\footnotesize{Pass@k vs. Sample Budget on FOLIO. (a) MoT-trained model with MoT sampling outperforms the base model (Gemma-2-9b-It) with SoT sampling. (b) Within the MoT-trained model, MoT sampling yields higher Pass@k than SoT sampling (NL\_CoT, Truth Table, Code).}}
  \label{fig:tts_with_MoT}
  \vspace{-1.5em}
\end{figure*}
We investigate how different single‑thought and MoT inference scale with an increased test‑time budget.
To do this, we first generate 128 responses from each model with each modality.
Then we evaluate two sampling strategies: 1) \textit{Single-Thought Sampling}: We randomly select \(k\) responses from the 128 generated responses. and 2) \textit{MoT Sampling}: Assuming there are \(N_T\) reasoning modalities, we sample \(\frac{k}{N_T}\) responses from each modality (so that the total number of responses is \(k\)). We choose $k$ ranging from 3 to 24 and have 10 runs for each setting.

\paragraph{MoT framework vs. Single-thought Baseline} We compare our Gemma-2-9b-It-MoT with Gemma-2-9b-It. Figure \ref{fig:tts_with_MoT} (a) shows our MoT model with MoT sampling consistently outperforms Gemma-2-9b-It with single-thought sampling. 
When the sample budget is less than 20, the performance gap is significant. 
It suggests that our MoT approach significantly increases the response diversity, leading to a more efficient use of inference compute. We observe a consistent phenomenon in terms of averaged accuracy (Appendix \ref{subsec:tts}, Figure \ref{fig:tts_vote_with_MoT}).

\paragraph{Comparison of different modalities} We further plot the scaling curves of our MoT model (based on Gemma-2-9B-It) under three reasoning modalities in Figure \ref{fig:tts_with_MoT} (b).
Here are insights: 1) While NL significantly outperforms the truth-table paradigm at low 
$k$, their theoretical upper bounds converge as 
$k$ increases; 2) The code paradigm exhibits the lowest upper bound among the three; 3) Across all values of 
$k$, our MoT framework consistently achieves the highest pass@k and attains the largest upper bound, indicating the largest potential of MoT trained models in test-time scaling.

\section{Further Analysis}
\label{sec:further_analysis}
\pgfplotsset{compat=1.18} 
\usetikzlibrary{calc}
\usepgfplotslibrary{groupplots}
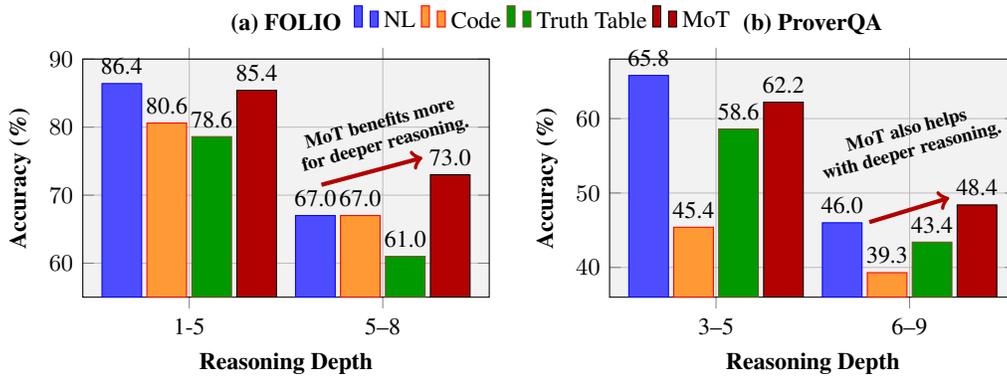
\begin{figure*}[t!]
\vspace{-2em}
\centering
\begin{tikzpicture}
\begin{groupplot}[
    group style={
        group size=2 by 1,
        horizontal sep=1.6cm
    },
    ybar,
    height=5cm,
    axis background/.style={fill=gray!10},
    enlarge x limits=0.46,
    grid=major,
    tick label style={font=\small},
    label style={font=\small},
    nodes near coords,
    every node near coord/.append style={
            font=\small,
            black
        },
        point meta=explicit symbolic,
    legend style={
        at={(0.33,1.25)},
        legend columns=4,
        draw=none,
        font=\small
    }
]

\nextgroupplot[
    title={\textbf{(a) FOLIO}},
    width=0.5\textwidth,
    height=0.34\textwidth,
    ymin=55, ymax=90,
    enlarge x limits=0.55,
    bar width=15pt,
    xlabel={\textbf{Reasoning Depth}},
    ylabel={\textbf{Accuracy (\%)}},
    symbolic x coords={1-5, 5--8},
    xtick=data
]
    \addplot+[fill=blue!70] coordinates {(1-5,86.4) [86.4] (5--8,67.0) [67.0]};
    \addplot+[fill=orange!80] coordinates {(1-5,80.6) [80.6] (5--8,67.0) [67.0]};
    \addplot+[fill=green!60!black] coordinates {(1-5,78.6) [78.6] (5--8,61.0) [61.0]};
    \addplot+[fill=red!70!black] coordinates {(1-5,85.4) [85.4] (5--8,73.0) [73.0]};

    \node[align=center, font=\small, text=black, rotate=15, scale=0.8] at (axis cs:5--8,79)
        {\textbf{MoT benefits more}\\\textbf{for deeper reasoning.}};

    \coordinate (motStartF) at (axis cs:1-5,75.4);
    \coordinate (motEndF) at (axis cs:5--8,72.0);
    \draw[->, ultra thick, red!70!black]
        ($(motStartF)+(50pt,-10pt)$) -- ($(motEndF)+(15pt,9pt)$);

\nextgroupplot[
    title={\textbf{(b) ProverQA}},
    width=0.5\textwidth,
    height=0.34\textwidth,
    ymin=36, ymax=68,
    enlarge x limits=0.55,
    bar width=15pt,
    xlabel={\textbf{Reasoning Depth}},
    ylabel={\textbf{Accuracy (\%)}},
    symbolic x coords={3--5, 6--9},
    xtick=data
]

    \addplot+[fill=blue!70] coordinates { (3--5,65.8) [65.8] (6--9,46.0)[46.0]};
    \addplot+[fill=orange!80] coordinates { (3--5,45.4) [45.4] (6--9,39.3) [39.3]};
    \addplot+[fill=green!60!black] coordinates {(3--5,58.6) [58.6] (6--9,43.4)[43.4]};
    \addplot+[fill=red!70!black] coordinates { (3--5,62.2) [62.2] (6--9,48.40) [48.4]};

    \node[align=center, font=\small, text=black, rotate=15, scale=0.8] at (axis cs:6--9,57)
        {\textbf{MoT also helps}\\\textbf{with deeper reasoning.}};

    \coordinate (motStartPQ) at (axis cs:3--5,50.7);
    \coordinate (motEndPQ) at (axis cs:6--9,60);
    \draw[->, ultra thick, red!70!black]
        ($(motStartPQ)+(58pt,-13pt)$) -- ($(motEndPQ)+(15pt,-30pt)$);

\legend{NL, Code, Truth Table, MoT}
\end{groupplot}
\end{tikzpicture}
\vspace{-1em}
\caption{\footnotesize{Performance comparison of different thought paradigms across reasoning depths. On FOLIO and ProverQA benchmarks, MoT inference exhibits better performance on difficult problems.}}
\vspace{-1em}
\label{fig:group-mot-depth}
\end{figure*}
\subsection{Mixture-of-Thought Inference Benefits More for Difficult Problems}
\label{sec:further_analysis_difficulty}
We further identify the types of problems that benefit most from the proposed MoT inference approach. Specifically, we focus on problem difficulty, which can be effectively measured by the depth of reasoning. We conduct analysis on FOLIO and ProverQA~\cite{qi2025large} benchmarks. Figure \ref{fig:group-mot-depth} shows the performance of our MoT model with different reasoning modalities across reasoning depths.

We can see that MoT inference benefits more in solving more difficult logical problems. Our final MoT model with MoT inference achieves an accuracy of 73.0\% on challenging logical tasks with reasoning depths ranging from 5 to 8, outperforming each modality by a significant margin, with an average improvement of 9\% points. However, such performance gains turn into slight degradation when dealing with easy problems. A similar phenomenon can be observed in ProverQA.

\subsection{Complementary, Uniqueness and Error Analysis}
\label{sec:further_analysis_error_overlap}
In this section, we quantify the complementary and uniqueness of our reasoning modalities and the training dynamics of our self‑evolving MoT training. We focus on three metrics:  
1) \emph{Unique coverage}, \emph{i.e.}, examples solved by exactly one modality;  2) \emph{Complementarity coverage}, \emph{i.e.}, examples solved by at least two modalities; and 3)  \emph{Oracle upper bound}, \emph{i.e.}, examples solved by at least one modality. 

Figure \ref{fig:motivation}(a),(b) shows each modality’s solve rate and oracle upper bound on ProofWriter and FOLIO. We further give a detailed unique and complementarity coverage and oracle upper bound in Table \ref{tab:overlap_across_datasets} in the Appendix.  
First, although our approach slightly reduces unique coverage compared to the baseline, both methods still achieve strong performance in this metric. Second, in terms of complementarity, our method increases the number of examples solved by multiple modalities—particularly on ProofWriter—demonstrating enhanced synergy. Third, by incorporating the truth‐table paradigm alongside Code and NL, our model attains a higher oracle upper bound than prior work using only Code+NL, underscoring the benefit and necessity of the truth‐table paradigm.

\label{sec:error_analysis}
\paragraph{Bottleneck of NL reasoning modality. }
We perform a human evaluation of model outputs generated by natural language reasoning on the FOLIO dataset. We identify two major error patterns in the incorrectly solved cases: 1) failure to consider multiple cases when handling disjunction operations, such as "either/or"; 2) failure to utilize the transposition inference rule. For example, given A → B, the model might sometimes incorrectly produce ¬A → ¬B. Motivated by these observations and error types identified in prior work~\cite{han-etal-2024-folio, olausson2023linc}, we define four error categories: (i) invalid converse; (ii) missing branch; (iii) factual misquote; and (iv) incorporation of commonsense knowledge and design an automatic pipeline to assess model rationales. Figure \ref{fig:motivation}(c) presents the results, showing that invalid converse and missing-branch errors together account for nearly 66\% of all errors. These findings further underscore the value of introducing the Truth Table thought paradigm.

\paragraph{Scenarios that Truth Table excels in.}
We manually analyze all 13 examples (Table~\ref{tab:overlap_across_datasets}) that were solved only using the truth table paradigm and find that 1) 5 out of 13 problems require transposition; 2) 5 out of 13 problems contain disjunction or similar operations (e.g., 'Rock can fly, or is a bird, or cannot breathe') and 3) 2 out of 13 problems contain both. This indicates that Truth Table may indeed complement the NL paradigm to some extent. We give two examples in Appendix \ref{subsec:example_by_tt}.

\vspace{-5pt}
\section{Related Work}
\vspace{-5pt}
\paragraph{LLMs for Symbolic Reasoning.}
Prior work has explored adapting LLMs to symbolic reasoning.
One common approach treats LLMs as nl-to-fol translators, and then use an external symbolic prover to derive the final answer~\cite{pan2023logic, olausson2023linc, callewaert2025verus, ryu2025divide}. While effective, this pipeline largely bypasses the model’s internal reasoning capabilities, which our work seeks to fully leverage. To alleviate this problem, another line of work seeks to directly leverage LLMs’ reasoning ability via CoT prompting~\cite{wei2022chain}. However, natural language remains inherently flexible and sometimes insufficient for structured reasoning. To bridge the gap between flexibility and formal rigor, recent work has explored combining natural and symbolic reasoning~\cite{xu-etal-2024-faithful, liu2024logic, Xiong2024HYBRIDMINDMS}. These approaches often either rely on a primary reasoning modality (e.g., symbolic or NL), augmented with auxiliary signals from other representations~\cite{xu-etal-2024-faithful, liu2024logic} or select one from multiple reasoning modalities~\cite{Xiong2024HYBRIDMINDMS} at inference time. In contrast, our work 1) explicitly defines three kinds of reasoning paradigms covering natural language, symbolic and code-based reasoning. 
2) goes beyond modality selection by jointly learning and inferring with all modalities, via a self-evolving MoT training and inference framework.
\paragraph{Encouraging Diverse Thinking in Chain-of-Thoughts.}
Previous work diversifies the CoT to further improve reasoning performance. A common strategy is to sample multiple outputs with higher temperatures~\cite{wang2022self,brown2024large}, but this cannot guarantee true diversity~\cite{wang2025typedthinker}. 
To address this, some work uses varied prompts—by task type~\cite{wang2025typedthinker}, difficulty~\cite{li2022competition}, or strategy~\cite{li2023think,he2024exploring}—and agent‑based prompting via multi‑agent debate~\cite{liang2023encouraging,Hegazy2024DiversityOT} or self‑reflection~\cite{zhang-etal-2024-self-contrast} to elicit diverse CoTs.
These methods diversify within one modality (NL or code). 
In contrast, we systematically introduce modality‑level diversity—truth table, natural language, and code reasoning—which better aligns with the structural requirements of symbolic tasks and complements existing approaches. Recent work has also explored training smaller models on diverse CoTs generated by large LLMs~\cite{ho2022large, Puerto2024FineTuningWD}, though these approaches are limited to single-modality supervision and rely on external teacher models. In contrast, our method introduces modality-level diversity and requires no external supervision. We demonstrate that inter-modality diversity yields greater benefits for self-training than intra-modality diversity in Appendix \ref{subsec:diverse_SoT_vs_MoT}.  Concurrent to our work, Chain-of-Reasoning (CoR)~\cite{yu2025chain} also explores diverse thinking with different reasoning modalities. However, CoR explores sequential collaboration among reasoning modalities and focuses on mathematical tasks, which is quite different from our work. In contrast, we differ in three key aspects: 1) We focus on logical reasoning tasks and proposed a truth table-based reasoning; 2) We develop a Mixture-of-Thought Self-Evolving Training algorithm to bootstrap the model’s reasoning capabilities; and 3) We exploit parallel synergy among reasoning modalities during both training and inference.

\paragraph{Self-evolving training.}
Self-evolving training techniques have been widely adopted to improve reasoning ability in LLMs, especially when there is lack of reasoning trajectories. Notably, \citet{zelikman2022star} propose a bootstrapping framework that iteratively generates and verifies reasoning trajectories based on the derived final answer, then fine-tunes the model on these self-labeled examples to improve reasoning performance with minimal human supervision. Following this idea, several works adapt self-evolving training to a wider range of tasks~\cite{hosseini2024v,Xiong2024HYBRIDMINDMS,zelikman2024self,wang2024self,lin2024lean,zelikman2024quiet,poesia2024certified,guan2025rstar,li2025start}. Additionally, researchers also explore improving the high-quality of rationales during STaR algorithm~\cite{poesia2024certified,guan2025rstar, peng2025regenesis}, incorporating techniques such as formal verification, monte carlo tree search, and abstract-to-concrete prompting. While previous work primarily focuses on generating higher-quality reasoning paths within a single modality, our work explores a complementary direction: how to jointly evolve and coordinate reasoning across multiple thought paradigms.

\section{Conclusion}

We presented \textbf{Mixture-of-Thought} (MoT), a unified framework for improving logical reasoning by enabling LLMs to reason through natural language, code-based, and symbolic (truth table) paradigms within a single system. Unlike previous work, our approach combines a self-evolving training process that fosters cross-paradigm synergy with an inference-time voting mechanism that aggregates complementary reasoning strategies. Extensive experiments on two challenging logical reasoning benchmarks, FOLIO and ProofWriter, demonstrate that MoT substantially outperforms strong baselines, particularly on complex, high-depth problems. 
\bibliographystyle{unsrtnat}
\bibliography{neurips_2024}

\medskip

\newpage
\appendix

\section{Broader Impact}
\label{app:broader_impact}
Our Mixture-of-Thought (MoT) framework is designed to improve logical reasoning by integrating multiple complementary modalities (natural language, code, truth tables). We do not foresee any direct societal harms specific to this method. Nevertheless, we offer the following considerations:
\begingroup
\setlist[itemize]{leftmargin=10pt, itemsep=2pt, topsep=0pt, partopsep=0pt, parsep=0pt}
\begin{itemize}
    \item Positive Impacts: 1) Model Efficiency: By enabling smaller, open‐source models (e.g., 9B parameters) to match or approach the reasoning performance of much larger closed‐source systems, MoT can lower computational and financial barriers for AI; 2)Cross‐disciplinary Integration: Our modular design was inspired by human cognition. It will foster interdisciplinary research and AI reasoning across fields.  
    \item Potential Risks: As a LLM-based approach, we encourage users to use with caution.
\end{itemize}
\endgroup

\section{Limitations and Future Work}
\label{sec:limitation}
While our Mixture-of-Thought (MoT) framework demonstrates strong performance on logical reasoning tasks, we have not evaluated its effectiveness on other types of reasoning tasks, such as mathematical or commonsense reasoning. Additionally, our test-time scaling experiments suggest promising directions—such as dynamic mixture-of-thought sampling under budget constraints—but our current work still has not fully explored the benefits of complementary reasoning modalities. Further exploring these aspects could be important to further push the performance boundary of open-source models on reasoning. Finally, Truth Table based reasoning still have room to improve as currently we only employ LLMs to generate Truth Table. 

We plan to further explore them in the three aspects:
\begingroup
\setlist[itemize]{leftmargin=10pt, itemsep=2pt, topsep=0pt, partopsep=0pt, parsep=0pt}
\begin{itemize}
    \item Extended to boarder tasks: currently our work cannot directly applied to reasoning tasks out of logical reasoning. This is because the reasoning modality we define in our work is specific for logical reasoning, \emph{e.g.,} Truth Table. Therefore, we plan to define more general but complementary reasoning modality that can be applied to more general broader of reasoning tasks and further show how our MoT framework can further improve performance of reasoning tasks beyond logical reasoning.
    \item  Inference-time collaboration on the fly. An interesting question is \textit{How can we fully leverage the benefits of complementary reasoning modalities during inference?} Recent Long CoT with RL~\cite{guo2025deepseek} has deliver remarkable performance in mathematical reasoning. A natural idea is to taking those reasoning modalities as atomic steps and performs sequential test-time scaling where the model alternate among reasoning modalities.
    \item  Improved Truth Table Reasoning: Intuitively, the ultimate role of the Truth-Table modality is not only to produce direct logical inferences but also to verify the outputs of other modalities (e.g., natural-language CoT). To strengthen this verification capability, we plan to design a more dedicated Truth-Table reasoning module. Specifically, we will: 1) Develop a multi-step construction process that incrementally builds and prunes the truth table; 2) Integrate a refinement module that checks and corrects intermediate rows or predicates; and 3) thereby enable the model to generate higher-quality truth-table-based rationales for both standalone inference and cross-modal consistency checks.
\end{itemize}
\endgroup
\section{Detailed Experimental Settings}
\label{sec:Detailed_Experimental_Settings}
\subsection{Datasets}
In this work, we adopt three logical reasoning datasets: 1) FOLIO~\cite{han-etal-2024-folio}, 2) ProofWriter~\cite{tafjord-etal-2021-proofwriter}, and 3) ProverQA~\cite{qi2025large}, to evaluate the effectiveness of our MoT framework. 
\paragraph{FOLIO~\cite{han-etal-2024-folio}.} FOLIO provides both the training and validation subsets, consisting of 1003 and 203 samples, respectively. There are two subsets with different difficulties: 1) HybLogic: contains 100 complex logical problems (5 – 8 reasoning steps)  and 2) WikiLogic: contains 103 simper logical problems (1 – 5 reasoning steps). In this work, we sample 1000 training samples from the FOLIO training set as seed dataset for our self-evovling MoT training and evaluate both baselines and our trained model on the FOLIO validation set.
\paragraph{ProofWriter~\cite{tafjord-etal-2021-proofwriter}.} ProofWriter is a synthetic dataset designed for evaluating the logical reasoning abilities of language models. It consists of multiple subsets, each containing logical reasoning problems of varying reasoning depths—from depth 0 (direct inference) up to depth 5 (requiring multi-step logical deductions). Following \citet{pan2023logic}, we select the most challenging subset (reasoning depth 5) to construct our training and test data. Specifically, we sample 1,000 instances from the training set provided by \citet{pan2023logic} as our training data and adopt their original test set directly for fair evaluation.

\paragraph{ProverQA~\cite{qi2025large}.} ProverQA is a recently proposed logical reasoning benchmark, notable for its large scale, high quality, and diversity. It consists of three subsets, each corresponding to a different reasoning difficulty level (\emph{i.e.}, reasoning depth). We select these subsets to evaluate the performance of our MoT framework across varying levels of reasoning complexity.

\subsection{Training Details}
We conduct all experiments on 4 H100 GPUs with Alignment Handbook~\cite{Tunstall_The_Alignment_Handbook}. For each dataset, we sample 1,000 training examples and perform 2–3 rounds of self-evolving training. In each round, the model is fine-tuned for 2 epochs with a learning rate of 2e-5 and a batch size of 128. We do not perform hyperparameter tuning. Further tuning may lead to better performance. All experiments are run with a fixed seed, \emph{i.e.,} 42, for reproducibility. 

\subsection{Inference Details}
We employ vLLM~\cite{kwon2023efficient} for efficient inference. During trajectory collection, we generate 10 reasoning traces per example using temperature 1.0, max\_tokens 2048, and sampling count 10. To maximize coverage while ensuring quality, we retain only the first generated trace that passes our quality check. For evaluation, we set the temperature to 0.7 and max\_tokens to 2048. All experiments are run with a fixed seed, \emph{i.e.,} 42, for reproducibility. 

\section{Full Prompts for MoT}
\label{app:full_prompts}
The full prompts we utilized in this work are illustrated as follows:
\begin{promptbox}[Full prompt used for Mixture‑of‑Thought]
You are a rigorous and logically precise AI assistant. Your task is to answer a logical reasoning problem strictly following one of three modes, as explicitly specified in the input. Only one mode will be present in the input. Follow that mode exclusively.

- Code Mode (<code> ... <end_of_code> <answer> ... <end_of_answer>)
  - If the input contains <code>, translate the problem into Python code.
  - Execute the logic and derive the answer.

- Natural Language Chain-of-Thought Mode (<nl_cot> ... <end_of_nl_cot> <answer> ... <end_of_answer>)
  - If the input contains <nl_cot>, solve the problem step by step in natural language.

- Truth Table Mode (<truth_table> ... <end_of_truth_table> <answer> ... <end_of_answer>)
  - If the input contains <truth_table>, construct a truth table and derive the answer from it.

### Rules
- Only use the mode specified in the input. Do not switch modes.
- Generate output strictly in the specified mode and format, with no additional text.
- Enclose all reasoning strictly within the corresponding mode tags.
- The final answer must be strictly enclosed in <answer> ... <end_of_answer>.
- Do not provide any reasoning or explanations outside of the designated mode tags.

The following is the problem you need to solve.

<premises>
{premises}
</premises>

<conclusion>
{conclusion}
</conclusion>

<question>
Is the following statement true, false, or uncertain? {conclusion}
</question>

<options>
(A) True
(B) False
(C) Uncertain
</options>

<{tag}>
\end{promptbox}

\begin{promptbox}[Full prompt used for Error Detection]
"You must determine whether a rationale faithfully justifies the truth value of a conclusion given a set of premises.\n\n"
        "Faithful means all and only the steps actually used in deriving the conclusion:\n"
        "- are grounded in the given premises or prior derived steps,\n"
        "- apply valid inference rules (no illicit converse or contraposition),\n"
        "- cover every disjunction branch or quantifier case,\n"
        "- use no unstated assumptions, external knowledge, or background commonsense,\n"
        "- and correctly assess whether the conclusion is supported or contradicted by the premises.\n\n"
        "You must also diagnose where and how the rationale fails when it is unfaithful, allowing trivial unused remarks to be overridden.\n\n"
        
        "Error Types:\n"
        "- Missing Branch: Failing to exhaustively consider all branches of a disjunction, conditionals, or quantified cases.\n"
        "- Invalid Converse: Illicitly reversing the direction of a conditional (e.g., mistaking 'A → B' for 'B → A').\n"
        "- Commonsense Injection: Using external background knowledge or commonsense not entailed or implied by the premises.\n"
        "- Factual Misquote: Misrepresenting, distorting, or misquoting the explicit content of the premises.\n\n"
        
        "Input (JSON):\n"
        "{\n"
        '  "premises":   "<string>",\n'
        '  "conclusion": "<string>",\n'
        '  "rationale":  "<string>",\n'
        '  "label":      "<string>",\n'
        '  "predict":    "<string>"\n'
        "}\n\n"
        
        "Output (JSON):\n"
        "{\n"
        '  "faithful":         true | false,\n'
        '  "error_type":       "<missing branch | invalid converse | commonsense injection | factual misquote>",\n'
        '  "error_location":   "<e.g., Step 3, Clause 2>",\n'
        '  "override":         true | false,\n'
        '  "analysis":         "<brief summary explaining why the reasoning is faithful or unfaithful, citing specific logical failures>"\n'
        "}\n\n"
        
        "Notes:\n"
        "- If multiple error types apply, list them all separated by commas.\n"
        "- Always identify the first point in the rationale where the faithfulness failure occurs.\n"
        "- Be concise, precise, and consistent in your labeling.\n\n"
        "Input:\n"
\end{promptbox}

\section{Additional Experimental Results}
\subsection{Evaluating LLM Performance Across Reasoning Modalities on FOLIO and ProofWriter}
\label{subsec:detailed_baseline_MoT_results_across_modalities}
\begin{table}[ht]
\centering
\caption{Performance of three models across reasoning modalities on FOLIO and ProofWriter.}
\label{table:baseline_results}
\vskip 0.1in
\label{tab:modality-benchmark}
\resizebox{\textwidth}{!}{\begin{tabular}{l ccc ccc}
\toprule
\multirow{2}{*}{Model} 
  & \multicolumn{3}{c}{FOLIO} 
  & \multicolumn{3}{c}{ProofWriter} \\ 
\cmidrule(lr){2-4} \cmidrule(lr){5-7}
  & NL    & Code   & Truth Table  & NL      & Code     & Truth Table \\
\midrule
Gemma-2-2B-It        &  42.4 & 38.4 & 36.5 & 39.8 & 40.8 & 37.5 \\
\quad + MoT training & 61.1 (18.7$\uparrow$) &  61.1 (22.7$\uparrow$) & 58.6 (22.1$\uparrow$) & 62.7 (22.9$\uparrow$) & 61.7 (20.9$\uparrow$)& 60.2 (22.7$\uparrow$)\\ 
Gemma-2-9B-It        & 69.5 & 56.7 & 63.6 & 61.2 & 39.5 & 55.8\\
\quad + MoT training & 76.9 (7.4$\uparrow$)  & 73.9 (17.2$\uparrow$) & 70.0 (6.4$\uparrow$) & 68.5 (7.3$\uparrow$) & 69.5 (30.0$\uparrow$)& 66.7 (10.9$\uparrow$)\\ 
Qwen-2.5-7B-Instruct & 71.9 & 62.1 & 69.0 & 60.5 & 42.3 & 53.0 \\
\quad + MoT training & 75.9 (4.0$\uparrow$)& 68.5 (6.4$\uparrow$) & 71.9 (2.9$\uparrow$) & 69.2 (8.7$\uparrow$)& 66.7 (24.4$\uparrow$)& 64.3 (11.3$\uparrow$) \\ 
\bottomrule
\end{tabular}}
\end{table}

Table \ref{table:baseline_results} displays detailed results of baselines across reasoning modalities on FOLIO and ProofWriter. We can observe that LLMs owns uneven ability across these reasoning modalities. This also highlights the necessary of our self-evolving MoT training, which can equip LLMs with three complementary reasoning modalities. After self-evolving MoT training, all modalities show joint improvements. This effect is especially significant in smaller models, \emph{i.e.}, Gemma-2-2B-It achieves up to a more than 20\% increase in accuracy on average.

\subsection{Comparison with more Baselines on FOLIO and ProofWriter}
\label{subsec:more_baseline_results}
\begin{table}[ht]
\centering
\caption{Comparison with more baselines on FOLIO and ProofWriter}
\label{tab:more_baseline_comparsion}
\vskip 0.1in
\label{tab:more-baselines}
\resizebox{0.8\textwidth}{!}{\begin{tabular}{l l c c c}
\toprule
\textbf{Method} & \textbf{Base Model} & \textbf{FOLIO (Acc \%)} & \textbf{ProofWriter (Acc \%)} \\
\midrule
HybridMind~\cite{Xiong2024HYBRIDMINDMS}      & GPT-3.5 & 76.6 & – \\
LINC~\cite{olausson2023linc}                  & GPT-4 & 72.5 & – \\
Symbolic CoT~\cite{xu2024faithful}           & GPT-4 & 83.3 & 82.5  \\
Logic-of-Thoughts @ 5       & GPT-3.5  & 81.5 & 65.9  \\
Logic-of-Thoughts @ 5       & GPT-4  & 88.2 & 72.0  \\
MoT                   & Gemma-2-2b-It & 62.6 & 65.0 \\
MoT                   & Gemma-2-9b-It & 78.9 & 70.7 \\
MoT                   & Qwen2.5-7B-Instruct & 78.3 & 71.8 \\
\bottomrule
\end{tabular}}
\end{table}

Table \ref{tab:more-baselines} presents a comparison between our approach and prior state-of-the-art systems. It demonstrates that our open-source MoT models nearly match the performance of leading closed-source prompting methods (\emph{e.g.}, GPT-3.5 and GPT-4). This indicates that enabling LLMs to learn complementary reasoning modalities is a promising direction.

\begin{table}[ht]
\centering
\caption{Ablation studies on (1) policy strategy; and (2) mixing strategy.}
\vskip 0.1in
\label{tab:ablation_studies}
\resizebox{\textwidth}{!}{\begin{tabular}{llccccc}
\toprule
\multirow{2}{*}{\textbf{Ablation}} & \multirow{2}{*}{\textbf{Setting}} & \multicolumn{4}{c}{FOLIO Accuracy (\%)} \\
\cmidrule(r){3-6}
& &  NL & Code & Truth Table & MoT\\
\midrule
\multirow{2}{*}{1. Policy Strategy} 
  & Off-policy MoT               &  55.2 & 54.7 & 53.7 &  56.7   \\
  & On-policy MoT (default)      & \bf 61.1   & \bf 61.1  & \bf 58.6  & \bf 62.6    \\
\midrule
\multirow{2}{*}{2. Mixing Strategy} 
  & Random single-modality per question       & 49.8  & 50.3 & 48.3 & 53.7   \\
  & Direct mixing (default)      & \bf 61.1   & \bf 61.1  & \bf 58.6  & \bf 62.6    \\
\bottomrule
\end{tabular}}
\end{table}

\subsection{Ablation Studies}
\label{subsec:ablation_studies}
We perform ablation studies on three core components: 1) policy strategy, \emph{i.e.,} on-policy vs. off-policy~\cite{zelikman2022star} and 2) mixing approach, \emph{i.e.,} direct mixture vs. mixture by unique conclusion (randomly select single-modality per question). 

Table~\ref{tab:ablation_studies} reports FOLIO accuracies under each setting. We make two key observations:
\begingroup
\setlist[itemize]{leftmargin=10pt, itemsep=2pt, topsep=0pt, partopsep=0pt, parsep=0pt}
\begin{itemize}
  \item \textbf{On-policy training yields consistent gains.}  
    Switching from off-policy to on-policy increases single-modality CoT accuracy by approximately 5–6 pp (e.g., NL CoT from 55.2\% to 61.1\%) and raises MoT’s final accuracy from 56.7\% to 62.6\%. This demonstrates the importance of updating the model with its most recent outputs.

  \item \textbf{Direct mixing outperforms random single-modality sampling.}  
    Presenting all three modalities together boosts accuracy by 8–10 pp compared to randomly picking one modality per question (MoT: 62.6\% vs.\ 53.7\%). This indicates that joint exposure to multiple modalities provides stronger complementary signals than isolated examples.
\end{itemize}
\endgroup
\subsection{Impact of Quality of Initial Training Data: Distillation + Single-Modal Training vs. Raw Data + MoT Training}
\label{subsec:effect_data_quality}
\definecolor{darkgreen}{RGB}{0,100,0}
\definecolor{tableauBlue}{RGB}{31,119,180}
\definecolor{tableauOrange}{RGB}{255,127,14}
\definecolor{tableauGreen}{RGB}{44,160,44}
\definecolor{tableauRed}{RGB}{214,39,40}
\definecolor{tableauPurple}{RGB}{148,103,189}
\pgfplotsset{compat=1.18}
\usepgfplotslibrary{groupplots,fillbetween}
\pgfplotsset{every axis/.append style={font=\small}}

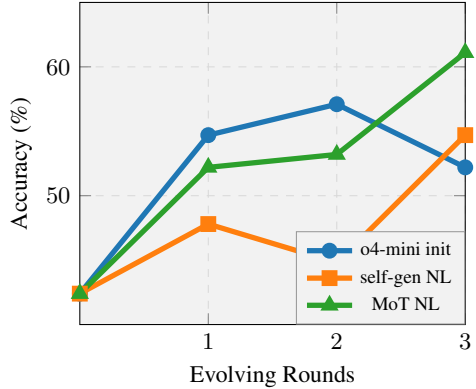
\begin{figure}[h]
  \centering
  \begin{tikzpicture}
    \begin{groupplot}[
      group style={group size=1 by 1},
      axis background/.style={fill=gray!10},
      width=0.48\textwidth,
      height=0.42\textwidth,
      xlabel={Evolving Rounds},
      ylabel={Accuracy (\%)},
      xmin=0, xmax=3,
      ymin=40, ymax=65,
      xtick={1,2,3},
      ytick={50,60,70,80},
      grid=both,
      grid style={dashed,gray!30},
      xlabel style={at={(axis description cs:0.5,-0.1)}, anchor=north},
      ylabel style={at={(axis description cs:-0.1,0.5)}, anchor=south},
      legend style={at={(axis description cs:1,0)}, anchor=south east, font=\small, draw=black!40, fill=gray!10}
    ]
      \nextgroupplot

      \addplot+[name path=o4mini, color=tableauBlue, line width=2pt, mark=*, mark size=2pt, mark options={fill=none}]
        coordinates {(0,42.4) (1,54.7) (2,57.1) (3,52.2)};
      \addlegendentry{\scalebox{0.8}{o4-mini init}}

      \addplot+[name path=selfNL, color=tableauOrange, line width=2pt, mark=square*, mark size=2pt, mark options={fill=none}]
        coordinates {(0,42.4) (1,47.8) (2,44.8) (3,54.7)};
      \addlegendentry{\scalebox{0.8}{self-gen NL}}

      \addplot+[name path=motNL, color=tableauGreen, line width=2pt, mark=triangle*, mark size=2pt, mark options={fill=none}]
        coordinates {(0,42.4) (1,52.2) (2,53.2) (3,61.1)};
      \addlegendentry{\scalebox{0.8}{MoT NL}}

    \end{groupplot}
  \end{tikzpicture}
  \caption{Accuracy (\%) over three self-evolving rounds on the FOLIO benchmark for: distilled NL-CoT (first-round only), raw NL-CoT (no distillation), and MoT (no distillation). The performance is evaluated with NL-based reasoning.}
  \label{fig:round-accuracy}
\end{figure}
Intuitively, the first‐round data are crucial and have a strong impact on the efficacy of self‐evolving training. Therefore, we are interested in the following question: \textit{Can self‐evolving single-thought training enhanced by first‐round distillation outperform our self‐evolving mixture-of-thought training without any distillation?} To answer this question, we compare the following settings: 1) Self-evolving single-thought (nl) training but with distillation data from o4-mini for first round training, which can provide a better initialization; 2) our MoT training without any distillation data; and 3) Self-evolving single-thought (nl) training without any distillation data. Figure \ref{fig:round-accuracy} displays the results of Gemma-2-2b-It on FOLIO benchmark. 

The key observations are: adding distillation data from stronger LLMs is beneficial for improving performance and convergence rate (blue line vs. orange line),  but still lags behind our self-evolving MoT training (blue line vs. green line). This suggests the advantages of our self-evolving MoT training: 1) It requires no reliance on stronger—often more expensive—LLMs; 2) It provides a higher upper bound accuracy.

\subsection{Finetuning with diverse Single-modality CoT vs. Finetuning with MoT}
\label{subsec:diverse_SoT_vs_MoT}
\citet{ho2022large, Puerto2024FineTuningWD} have explored that finetuning LLMs with diverse CoT can further improve the performance. A natural question then is: \textit{given a fixed budget of training examples, which strategy yields better results?  (1) self-training with $3N$ natural-language CoT samples, or (2) self-training with a total of $3N$ samples composed of $N$ examples from each of three modalities (NL, Code, Truth-Table)?}

We consider two settings to answer this question: 1) Self-evolving training with $3N$ natural-language CoT samples for \emph{2 epochs per round} over \emph{3 rounds}. We sample 10 reasoning traces per question with temperature of 1.0 and keep the 3 reasoning traces that satisfy our filtering criteria; 2) Self-evolving training with a total of $3N$ samples comprising $N$ examples from each of the three modalities (NL, Code, Truth-Table) for \emph{2 epochs per round} over \emph{3 rounds}. We evaluate those trained model with natural language modality on FOLIO dataset.

\begin{table}[ht]
\centering
\caption{Accuracy (\%) of Gemma-2-2b-It under three self-evolving regimes, with budgets of $N$ or $3N$ training samples. The accuracy is evaluated with NL-based reasoning on FOLIO benchmark. We can see self-evolving training with MoT achieves the best accuracy, demonstrating the benefit of modality-level diversity.}
\label{tab:finetune-ablation}
\vskip 0.1in
\begin{tabular}{llcccc}
\toprule
\# & \textbf{Setting} & \bf Training Samples & \textbf{Accuracy (\%)} \\
\midrule
1 & NL\_CoT  & $N$ 
    & 54.7 \\
2 & NL\_CoT  & 3$N$ 
    & 57.1 \\
3 &  MoT data & 3$N$  & 61.1 \\
\bottomrule
\end{tabular}
\end{table}

Table \ref{tab:finetune-ablation} shows the results. We can have the following observations: 1) finetuning with diverse NL CoT can indeed improve the performance (\#1 vs. \#2), which is consistent with findings from \citet{ho2022large, Puerto2024FineTuningWD}. 2) Finetuning with MoT data is more efficient than finetuning with same amount of diverse NL CoT data  (\#2 vs. \#3). This indicates that the diversity of single-modality CoT data obtained by sampling with high temperature is not sufficient. By contrast, our MoT data, which leverages the complementarity of truth table, code and nl, can produce more diversity, and therefore improve the training efficiency.

\subsection{Additional Results on Test-Time Scaling Across Reasoning Modalities}
\label{subsec:tts}
\paragraph{MoT With Different Thought Paradigms} 
Table~\ref{fig:tts_with_MoT} (b) illustrates the scaling behavior of our MoT model across different thought paradigms under varying sample budgets. We observe that code-based reasoning consistently lags behind all other paradigms, indicating its relatively poor performance and limited scalability. 

Another interesting phenomenon is that natural language-based reasoning achieves relatively strong performance when the sample budget is small (e.g., $k < 20$), outperforming the truth table-based paradigm in this regime. However, as the sample budget increases (e.g., $k > 20$), truth table reasoning begins to match even outperform NL-based reasoning—highlighting its greater potential when more inference resources are available. 

Notably, our MoT (ALL) approach offers a favorable trade-off between these two paradigms: it achieves strong performance under low-budget conditions, while delivering better performance when the sample budget is large. 

\paragraph{Accuracy vs. Sample Budget} \definecolor{darkgreen}{RGB}{0,100,0}
\definecolor{tableauBlue}{RGB}{31,119,180}
\definecolor{tableauOrange}{RGB}{255,127,14}
\definecolor{tableauGreen}{RGB}{44,160,44}
\definecolor{tableauRed}{RGB}{214,39,40}
\definecolor{tableauPurple}{RGB}{148,103,189}
\pgfplotsset{compat=1.18}
\usepgfplotslibrary{groupplots,fillbetween}
\pgfplotsset{every axis/.append style={font=\small}}

\begin{wrapfigure}[15]{r}{0.47\textwidth}
  \centering
  \vspace{-2em}
  \begin{tikzpicture}
    \begin{groupplot}[
      group style={group size=1 by 1, horizontal sep=1.5cm},
      axis background/.style={fill=gray!10},
      width=0.48\textwidth,
      height=0.42\textwidth,
      xlabel={Sample Budget},
      ylabel={Accuracy (\%)},
      xmin=0, xmax=28,
      ymin=40, ymax=85,
      grid=both,
      grid style={dashed,gray!30},
      xlabel style={at={(axis description cs:0.5,-0.1)}, anchor=north},
      ylabel style={at={(axis description cs:-0.1,0.5)}, anchor=south}
    ]
      \nextgroupplot[
        legend style={at={(axis description cs:1,0)}, anchor=south east, font=\small, draw=black!40, fill=gray!10}
      ]
      
      \addplot+[name path=base, color=blue, line width=2pt, mark=*, mark size=2pt, mark options={fill=none}]
        table [x index=0, y index=1] {data/Accuracy_vs_Sample_size/FOLIO_Baseline_NL.dat};
      \addlegendentry{\scalebox{0.8}{Gemma-2-9b-It (NL\_CoT)}}

      \addplot+[name path=base_tt, color=darkgreen!50, line width=2pt, mark=*, mark size=2pt, mark options={fill=none}]
        table [x index=0, y index=1] {data/Accuracy_vs_Sample_size/FOLIO_Baseline_TT.dat};
      \addlegendentry{\scalebox{0.8}{Gemma-2-9b-It (Truth Table)}}

      \addplot+[name path=base_code, color=tableauOrange!50, line width=2pt, mark=*, mark size=2pt, mark options={fill=none}]
        table [x index=0, y index=1] {data/Accuracy_vs_Sample_size/FOLIO_Baseline_Code.dat};
      \addlegendentry{\scalebox{0.8}{Gemma-2-9b-It (Code)}}


      \addplot+[name path=moth, color=red, line width=2pt, mark=*, mark size=2pt, mark options={fill=none}]
        table [x index=0, y index=1] {data/Accuracy_vs_Sample_size/FOLIO_MoT.dat};
      \addlegendentry{\scalebox{0.8}{Gemma-2-9b-It-MoT (MoT)}}

      \addplot+[name path=baseupper, draw=none, mark=none]
        table [x index=0, y expr=\thisrowno{1}+\thisrowno{2}] {data/Accuracy_vs_Sample_size/FOLIO_Baseline_NL.dat};
      \addplot+[name path=baselower, draw=none, mark=none]
        table [x index=0, y expr=\thisrowno{1}-\thisrowno{2}] {data/Accuracy_vs_Sample_size/FOLIO_Baseline_NL.dat};
      \addplot [blue!20, opacity=0.7, forget plot, on layer=axis background]
        fill between[of=baseupper and baselower];

    \addplot+[name path=basettupper, draw=none, mark=none]
        table [x index=0, y expr=\thisrowno{1}+\thisrowno{2}] {data/Accuracy_vs_Sample_size/FOLIO_Baseline_TT.dat};
      \addplot+[name path=basettlower, draw=none, mark=none]
        table [x index=0, y expr=\thisrowno{1}-\thisrowno{2}] {data/Accuracy_vs_Sample_size/FOLIO_Baseline_TT.dat};
      \addplot [darkgreen!20, opacity=0.7, forget plot, on layer=axis background]
        fill between[of=basettupper and basettlower];

    \addplot+[name path=basecodeupper, draw=none, mark=none]
        table [x index=0, y expr=\thisrowno{1}+\thisrowno{2}] {data/Accuracy_vs_Sample_size/FOLIO_Baseline_Code.dat};
      \addplot+[name path=basecodelower, draw=none, mark=none]
        table [x index=0, y expr=\thisrowno{1}-\thisrowno{2}] {data/Accuracy_vs_Sample_size/FOLIO_Baseline_Code.dat};
      \addplot [tableauOrange!30, opacity=0.7, forget plot, on layer=axis background]
        fill between[of=basecodeupper and basecodelower];

      
      \addplot+[name path=mothupper, draw=none, mark=none]
        table [x index=0, y expr=\thisrowno{1}+\thisrowno{2}] {data/Accuracy_vs_Sample_size/FOLIO_MoT.dat};
      \addplot+[name path=mothlower, draw=none, mark=none]
        table [x index=0, y expr=\thisrowno{1}-\thisrowno{2}] {data/Accuracy_vs_Sample_size/FOLIO_MoT.dat};
      \addplot [red!20, opacity=0.7, forget plot, on layer=axis background]
        fill between[of=mothupper and mothlower];

    \end{groupplot}
  \end{tikzpicture}
  \caption{Accuracy vs. Sample Budget for different modes}
  \label{fig:tts_vote_with_MoT}
\end{wrapfigure}
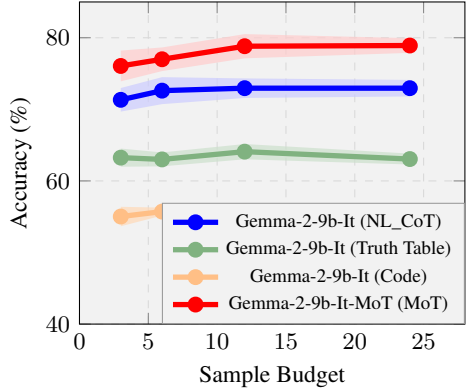 Figure~\ref{fig:tts_vote_with_MoT} presents accuracy-vs-sample-budget curves across different reasoning paradigms. We find that our MoT (ALL) model—trained and inferred under the mixture-of-thought setting—consistently achieves the highest accuracy, outperforming all other approaches regardless of budget size. Additionally, our MoT model can benefit better from increased sample budget compared wiht all other approaches. Among individual paradigms, NL-CoT performs best under majority voting, while truth table reasoning is more stable but shows limited improvement with increased budget. Code-based reasoning remains the least effective. These results reinforce the value of our MoT framework.


\subsection{Detailed Complementary, Uniqueness Analysis}
\label{app:complementary_uniqueness_analysis}
\begin{wraptable}[16]{r}{0.4\textwidth}
\vspace{-1.5em}
\setlength{\tabcolsep}{1.5pt}
\centering
\caption{Prediction Category Distribution on Two Benchmarks (Qwen-2.5-7B-Instruct vs Qwen-2.5-7B-Instruct-MoT).}
\label{tab:overlap_across_datasets}
\resizebox{0.4\textwidth}{!}{\begin{tabular}{lcccc}
\toprule
\multirow{2}{*}{\textbf{ Category}} & \multicolumn{2}{c}{\textbf{ProofWriter}} & \multicolumn{2}{c}{\textbf{FOLIO}} \\
\cmidrule(lr){2-3} \cmidrule(lr){4-5}
& Baseline & Our & Baseline & Our \\
\midrule
\multicolumn{5}{l}{\textit{Single-paradigm only}} \\
\quad Only NL correct & 109 & 55 & 18 & 8 \\
\quad Only Code correct & 40 & 32 & 5 & 6 \\
\quad Only TT correct & 66 & 33 & 11 & 13 \\
\midrule
\multicolumn{5}{l}{\textit{Pairwise overlap only}} \\
\quad NL $\cap$ Code only & 172 & 304 & 109 & 125 \\
\quad NL $\cap$ TT only & 210 & 289 & 117 & 125 \\
\quad Code $\cap$ TT only & 170 & 297 & 110 & 112 \\
\midrule
Code $\cup$ NL  & 445 & 511 & 163 & 168\\
Code $\cup$ NL $\cup$ TT & 511 & \textbf{544} & 174 & \textbf{181} \\
\bottomrule
\end{tabular}}
\end{wraptable}
Across both ProofWriter and FOLIO benchmarks, our Mixture‑of‑Thought (MoT) model shifts away from single‑paradigm reliance and toward multi‑paradigm collaboration. First, the number of examples solved exclusively by the NL paradigm drops by over 50\% (ProofWriter: from 109 to 55; FOLIO: from 18 to 8), and “Only TT correct” cases likewise decrease, indicating that MoT reduces brittle, single‑mode reasoning. Second, pairwise overlaps (NL $\cap$ Code, NL $\cap$ TT, Code $\cap$ TT) all increase substantially—NL $\cap$ Code on ProofWriter rises by 76\% (172 → 304), and similar gains appear on FOLIO—showing that MoT effectively combines different reasoning formats on the same instance. Finally, the overall coverage (Code $\cup$ NL $\cup$ TT) improves from 511 to 544 (+6.5\%) on ProofWriter and from 174 to 181 (+4\%) on FOLIO, demonstrating that MoT recovers difficult cases missed by the baseline. The consistent trends across two datasets confirm that encouraging multi‑paradigm synergy yields more robust and comprehensive logical reasoning performance.



\section{More Examples Generated By Our MoT}
\label{sec:more_examples_model_outputs}
In this section, we show more examples generated by our MoT with different thought paradigms.
\subsection{Natural Language based Reasoning}
\begin{promptbox}[Examples of NL-based Reasoning generated by our models.]
The following is the problem you need to solve.

<premises>
All rabbits that can be spotted near the campus are cute. 
Some turtles can be spotted near the campus. 
The only animals that can be spotted near the campus are rabbits and squirrels.
If something is skittish, then it is not calm.
All the squirrels that can be spotted near the campus are skittish.
Rockie can be spotted near the campus, and it is calm.
</premises>

<conclusion>
Rockie is a turtle or cute.
</conclusion>

<question>
Is the following statement true, false, or uncertain? Rockie is a turtle or cute.
</question>

<options>
(A) True
(B) False
(C) Uncertain
</options>

<nl_cot>
Step 1: We know from the premise "The only animals that can be spotted near the campus are rabbits and squirrels." that Rockie, being spotted near the campus, must be either a rabbit or a squirrel. 
Step 2: We also know from the premise "All rabbits that can be spotted near the campus are cute" that if Rockie is a rabbit, then Rockie is cute. 
Step 3: The premise "All the squirrels that can be spotted near the campus are skittish" and "If something is skittish, then it is not calm" tell us that if Rockie is a squirrel, then Rockie is not calm. 
Step 4: The premise "Rockie can be spotted near the campus, and it is calm" contradicts the conclusion from Step 3. This means Rockie cannot be a squirrel. 
Step 5: Therefore, Rockie must be a rabbit. 
Step 6: Since Rockie is a rabbit and all rabbits spotted near the campus are cute, we can conclude that Rockie is cute.

<end_of_nl_cot>
<answer>
The final answer is (A).
<end_of_answer>
\end{promptbox}

\subsection{Code based Reasoning}
\begin{promptbox}[Examples of Code-based Reasoning generated by our models.]
The following is the problem you need to solve.

<premises>
Everyone at the mixer is a Grand Slam champion or an Oscar-nominated actor.
Every Grand Slam champion at the mixer is a professional tennis player.
All Oscar-nominated actors at the mixer are celebrities.
All professional tennis players at the mixer are athletes.
If a person at the mixer is a celebrity, then they are well paid.
If a person at the mixer is an athlete, then they are famous.
All well-paid people at the mixer live in tax havens.
Djokovic is at the mixer: if Djokovic is a famous athlete, then Djokovic is well-paid.
</premises>

<conclusion>
Djokovic does not live in a tax haven.
</conclusion>

<question>
Is the following statement true, false, or uncertain? Djokovic does not live in a tax haven.
</question>

<options>
(A) True
(B) False
(C) Uncertain
</options>

<code>
from typing import Optional

class Person:
    def __init__(self,
                 is_grand_slam_champion: Optional[bool] = None,
                 is_oscar_nominated_actor: Optional[bool] = None,
                 is_professional_tennis_player: Optional[bool] = None,
                 is_celebrity: Optional[bool] = None,
                 is_athlete: Optional[bool] = None,
                 is_well_paid: Optional[bool] = None,
                 lives_in_tax_haven: Optional[bool] = None,
                 is_famous: Optional[bool] = None,
                 name: str = "Djokovic"):
        self.is_grand_slam_champion = is_grand_slam_champion
        self.is_oscar_nominated_actor = is_oscar_nominated_actor
        self.is_professional_tennis_player = is_professional_tennis_player
        self.is_celebrity = is_celebrity
        self.is_athlete = is_athlete
        self.is_well_paid = is_well_paid
        self.lives_in_tax_haven = lives_in_tax_haven
        self.is_famous = is_famous

def apply_premises(person: Person) -> bool:
    changed = False

    # Premise 1
    if person.is_grand_slam_champion is None and person.is_oscar_nominated_actor is None:
        person.is_grand_slam_champion = True
        changed = True

    # Premise 2
    if person.is_grand_slam_champion is True and person.is_professional_tennis_player is None:
        person.is_professional_tennis_player = True
        changed = True

    # Premise 3
    if person.is_oscar_nominated_actor is True and person.is_celebrity is None:
        person.is_celebrity = True
        changed = True

    # Premise 4
    if person.is_professional_tennis_player is True and person.is_athlete is None:
        person.is_athlete = True
        changed = True

    # Premise 5
    if person.is_celebrity is True and person.is_well_paid is None:
        person.is_well_paid = True
        changed = True

    # Premise 6
    if person.is_athlete is True and person.is_famous is None:
        person.is_famous = True
        changed = True

    # Premise 7
    if person.is_well_paid is True and person.lives_in_tax_haven is None:
        person.lives_in_tax_haven = True
        changed = True

    # Premise 8
    if person.name == "Djokovic" and person.is_famous is True and person.is_well_paid is None:
        person.is_well_paid = True
        changed = True

    return changed

def run_inference(person: Person):
    while apply_premises(person):
        pass

def check_conclusion(person: Person) -> str:
    run_inference(person)
    if person.name == "Djokovic" and person.lives_in_tax_haven is True:
        return "False"
    else:
        return "True"

def func():
    person = Person(
        is_grand_slam_champion=None,
        is_oscar_nominated_actor=None,
        is_professional_tennis_player=None,
        is_celebrity=None,
        is_athlete=None,
        is_well_paid=None,
        lives_in_tax_haven=None,
        is_famous=None,
        name="Djokovic"
    )
    return check_conclusion(person)

if __name__ == '__main__':
    result = func()
    print("Conclusion: Djokovic does not live in a tax haven?", result)
<end_of_code>
<answer>
The final answer is (B).
<end_of_answer>
\end{promptbox}
\subsection{Truth Table based Reasoning}
\label{subsec:example_tt}

\begin{promptbox}[Examples of Truth Table-based Reasoning generated by our models.]
The following is the problem you need to solve.

<premises>
People in this club who perform in school talent shows often attend and are very engaged with school events.
People in this club either perform in school talent shows often or are inactive and disinterested community members.
People in this club who chaperone high school dances are not students who attend the school.
All people in this club who are inactive and disinterested members of their community chaperone high school dances.
All young children and teenagers in this club who wish to further their academic careers and educational opportunities are students who attend the school.
Bonnie is in this club and she either both attends and is very engaged with school events and is a student who attends the school or is not someone who both attends and is very engaged with school events and is not a student who attends the school.
</premises>

<conclusion>
Bonnie performs in school talent shows often.
</conclusion>

<question>
Is the following statement true, false, or uncertain? Bonnie performs in school talent shows often.
</question>

<options>
(A) True
(B) False
(C) Uncertain
</options>

<truth_table>
Let:
    T: Bonnie performs in school talent shows often.
    E: Bonnie attends and is very engaged with school events.
    S: Bonnie is a student who attends the school.
    I: Bonnie is an inactive and disinterested community member.
    C: Bonnie chaperones high school dances.

Premises:
1. E → S
2. T ∨ I
3. C → ¬S
4. I → C
5. S → E
6. (E ∧ S) ∨ (¬E ∧ ¬S)

Logical chain (assuming T = False):
    From (2): T = False -> I = True
    From (4): I = True -> C = True
    From (3): C = True -> ¬S = True -> S = False
    From (6): S = False -> ¬E = True -> E = False
    From (1): E = False -> E→S holds vacuously
    -> No contradiction if T = False

Logical chain (assuming T = True):
    From (2): T = True -> I = False
    From (4): I = False -> C = False
    From (3): C = False -> ¬S = False -> S = True
    From (5): S = True -> E = True
    From (6): E = True ∧ S = True -> (E∧S) = True
    -> No contradiction if T = True

Truth Table (remaining consistent rows):
| T     | E     | S     | I     | C     |
|-------|-------|-------|-------|-------|
| True  | True  | True  | False | False |
| False | False | False | True  | True  |

<end_of_truth_table>
<answer>
The final answer is (C).
<end_of_answer>
\end{promptbox}

\begin{promptbox}[Examples of Truth Table-based Reasoning generated by our models.]
The following is the problem you need to solve.

<premises>
All fir trees are evergreens.
Some objects of worship are fir trees.
</premises>

<conclusion>
Some evergreens are not objects of worship.
</conclusion>

<question>
Is the following statement true, false, or uncertain? Some evergreens are not objects of worship.
</question>

<options>
(A) True
(B) False
(C) Uncertain
</options>

<truth_table>
Let:
- F: An object is a fir tree.
- E: An object is an evergreen.
- W: An object is an object of worship.

Premises:
1. All fir trees are evergreens.  
   F → E
2. Some objects of worship are fir trees.  
   ∃x (W(x) ∧ F(x))

Conclusion to Evaluate:
Some evergreens are not objects of worship.  
∃x (E(x) ∧ ¬W(x))

Logical Chain:
1. From (1), F → E.
2. From (2), there exists some object a such that W(a) ∧ F(a).
3. Since F(a) and F → E, it follows that E(a).
4. Therefore, we have E(a) and W(a).
5. We need ∃x (E(x) ∧ ¬W(x)).
6. The premises do not provide any information about whether there exists an evergreen that is not an object of worship.
7. Therefore, it is uncertain whether some evergreens are not objects of worship.

Truth Table (Key Possibilities):

| F | E    | W    | E ∧ ¬W |
| True | True | True | False  |
| True | True | False| True   |
| False| True | True | False  |
| False| True | False| True   |

Rows omitted: any with E= False (irrelevant to E∧¬W) or with F= True and E= False (violates F→E).

<end_of_truth_table>
<answer>
The final answer is (C).
<end_of_answer>

\end{promptbox}

\section{Error Analysis}
\subsection{Examples of Error Type}
\label{subsec:example_error_types}

\begin{promptbox}[Example 1 of Missing Branches]

<premises>
All employees who schedule a meeting with their customers will go to the company building today. 
Everyone who has lunch in the company building schedules meetings with their customers. 
Employees have lunch either in the company building or at home.
If an employee has lunch at home, they are working remotely from home.
All employees who are in other countries work remotely from home. 
No managers work remotely from home. 
James will appear in the company today if and only if he is a manager.
</premises>

<conclusion>
James has lunch in the company.
</conclusion>

<question>
Is the following statement true, false, or uncertain? James has lunch in the company.
</question>

<options>
(A) True
(B) False
(C) Uncertain
</options>

<nl_cot>
Step 1: From the premise \"All employees who schedule a meeting with their customers will go to the company building today,\" we know that if an employee schedules a meeting with their customers, they will be in the company building today.

Step 2: From the premise \"Everyone who has lunch in the company building schedules meetings with their customers,\" we can infer that if an employee has lunch in the company building, they must schedule a meeting with their customers. 

Step 3: From the premise \"Employees have lunch either in the company building or at home,\" we know that an employee can only choose one of these two options for lunch.

Step 4: From the premise \"If an employee has lunch at home, they are working remotely from home,\" we can deduce that having lunch at home means they are not in the company building and are working remotely.

Step 5: From the premise \"All employees who are in other countries work remotely from home,\" we know that employees in other countries do not have lunch in the company building.

Step 6: From the premise \"No managers work remotely from home,\" we know that managers do not have lunch at home and thus do not work remotely.

Step 7: From the premise \"James will appear in the company today if and only if he is a manager,\" we can infer that if James is not a manager, he will not appear in the company today. 

Step 8: Combining the information from steps 1-7, we can conclude that if James is a manager, he will be in the company today. Since James will only be in the company today if and only if he is a manager, and managers do not work remotely, James cannot be having lunch at home.
(*@\textcolor{red}{$\Rightarrow$Incorrect. It only considers the case where James is a manger, missing another branch: James is not a manger.  }@*)

Step 9: Therefore, if James is not having lunch at home, he must be having lunch in the company building, and by step 2, he must be scheduling meetings with his customers, which means he will be in the company building today.

<end_of_nl_cot>
<answer>
The final answer is (A).
<end_of_answer>

<Label>
(C) Uncertain
</Label>
\end{promptbox}

\begin{promptbox}[Example 2 of Both Missing Branches and Invalid Converse  ]

<premises>
People in this club who perform in school talent shows often attend and are very engaged with school events.
People in this club either perform in school talent shows often or are inactive and disinterested community members.
People in this club who chaperone high school dances are not students who attend the school.
All people in this club who are inactive and disinterested members of their community chaperone high school dances.
All young children and teenagers in this club who wish to further their academic careers and educational opportunities are students who attend the school. 
Bonnie is in this club and she either both attends and is very engaged with school events and is a student who attends the school or is not someone who both attends and is very engaged with school events and is not a student who attends the school.
</premises>

<conclusion>
Bonnie performs in school talent shows often.
</conclusion>

<question>
Is the following statement true, false, or uncertain? Bonnie performs in school talent shows often.
</question>

<options>
(A) True
(B) False
(C) Uncertain
</options>

<nl_cot>
Step 1: Analyze the given premises and the statement about Bonnie.
    - Premise 1: People in this club who perform in school talent shows often attend and are very engaged with school events.
    - Premise 2: People in this club either perform in school talent shows often or are inactive and disinterested community members.
    - Premise 3: People in this club who chaperone high school dances are not students who attend the school.
    - Premise 4: All people in this club who are inactive and disinterested members of their community chaperone high school dances.
    - Premise 5: All young children and teenagers in this club who wish to further their academic careers and educational opportunities are students who attend the school.
    - Bonnie's statement: Bonnie is in this club and she either both attends and is very engaged with school events and is a student who attends the school or is not someone who both attends and is very engaged with school events and is not a student who attends the school.
Step 2: Determine Bonnie's status in the club.
    - According to Bonnie's statement, she is either a very engaged and attending member or not such a member. This statement is a disjunction (or) of two conditions, which means Bonnie can only fit into one of the two categories described.
Step 3: Consider the implications for Bonnie based on her status in the club.
    - If Bonnie is a very engaged and attending member, she must perform in school talent shows often (Premise 1).
    (*@\textcolor{red}{$\Rightarrow$Incorrect (Invalid Converse). It incorrectly utilizes "Bonnie is a very engaged and attending member -> perform".  }@*)
    - If Bonnie is not a very engaged and attending member, she must be an inactive and disinterested community member (Premise 2). Inactive and disinterested members must chaperone high school dances (Premise 4), which means they cannot be students who attend the school (Premise 3).
    (*@\textcolor{red}{$\Rightarrow$Incorrect (Missing Branches). It fails to enumerate all possible cases stated in Premise.  }@*)
Step 4: Evaluate the given statement.
    - The statement claims that Bonnie performs in school talent shows often.
    - Given the conditions and implications, if Bonnie is not a very engaged and attending member, she must be an inactive and disinterested member, which means she would be a chaperone and not a student, contradicting Premise 5. Therefore, Bonnie must be a very engaged and attending member.
Step 5: Conclude based on the analysis.
    - Since Bonnie must be a very engaged and attending member, she performs in school talent shows often.
<end_of_nl_cot>
<answer>
The final answer is (A).
<end_of_answer>


<Label>
(C) Uncertain
</Label>
\end{promptbox}

\begin{promptbox}[Example 3 of Missing Branches  ]

<premises>
All of Zaha Hadid's design styles that Max adores have interesting geometries.
No brutalist buildings that Max adores have interesting geometries.
Every style that Max adores is either Zaha Hadid's design style or Kelly Wearstler's design style. 
All of Kelly Wearstler's design styles that Max adores are evocative.
All of Kelly Wearstler's design styles that Max adores are dreamy.
If a design by Max that he adores has interesting geometries, then the design is a brutalist building and evocative.
</premises>

<conclusion>
A design by Max is evocative and dreamy.
</conclusion>

<question>
Is the following statement true, false, or uncertain? A design by Max is evocative and dreamy.
</question>

<options>
(A) True
(B) False
(C) Uncertain
</options>

<nl_cot>
Step 1: Analyze the premise \"All of Zaha Hadid's design styles that Max adores have interesting geometries.\" This means if a design by Max adores is a Zaha Hadid design, then it has interesting geometries.

Step 2: Analyze the premise \"No brutalist buildings that Max adores have interesting geometries.\" This means if a design by Max adores is a brutalist building, it does not have interesting geometries.

Step 3: Analyze the premise \"Every style that Max adores is either Zaha Hadid's design style or Kelly Wearstler's design style.\" This means all designs by Max adores must be either Zaha Hadid's or Kelly Wearstler's.

Step 4: Analyze the premise \"All of Kelly Wearstler's design styles that Max adores are evocative.\" This means if a design by Max adores is a Kelly Wearstler design, then it is evocative.

Step 5: Analyze the premise \"All of Kelly Wearstler's design styles that Max adores are dreamy.\" This means if a design by Max adores is a Kelly Wearstler design, then it is dreamy.

Step 6: Analyze the premise \"If a design by Max that he adores has interesting geometries, then the design is a brutalist building and evocative.\" This means if a design by Max adores has interesting geometries, then it is both a brutalist building and evocative.

Step 7: From Step 1 and Step 6, if a design by Max adores is a Zaha Hadid design and has interesting geometries, it must also be a brutalist building and evocative.

Step 8: From Step 2, if a design by Max adores is a brutalist building, it does not have interesting geometries. So, a design by Max adores cannot be both a brutalist building and have interesting geometries.

Step 9: Since a design by Max adores cannot have both interesting geometries (which would make it a brutalist and evocative) and not have interesting geometries (which would make it not a brutalist), the design by Max adores cannot have interesting geometries.
(*@\textcolor{red}{$\Rightarrow$Incorrect (Missing Branches). It fails to enumerate all possible cases stated in Premise 3.  }@*)

Step 10: Since the design by Max adores does not have interesting geometries, according to Step 4 and Step 5, it cannot be a Kelly Wearstler design. Therefore, it must be a Zaha Hadid design.

Step 11: Since the design by Max adores is a Zaha Hadid design, it must have interesting geometries (from Step 1), but this contradicts the conclusion that it cannot have interesting geometries (from Step 8).

Step 12: Given the premises, it is impossible for a design by Max adores to be both evocative and dreamy, as it would require having interesting geometries and not having them at the same time.
<end_of_nl_cot>
<answer>
The final answer is (B).
<end_of_answer>



<Label>
(A) True
</Label>
\end{promptbox}

\subsection{Examples uniquely solved by Truth Table}
\label{subsec:example_by_tt}

\begin{promptbox}[Examples uniquely solved by Truth Table]
<premises>
"Stranger Things" is a popular Netflix show.
If a Netflix show is popular, Karen will binge-watch it.
If and only if Karen binge-watches a Netflix show, she will download it.
Karen does not download "Black Mirror."
"Black Mirror" is a Netflix show.
If Karen binge-watches a Netflix show, she will share it with Lisa.
</premises>

<conclusion>
"Black Mirror" is popular.
</conclusion>

<question>
Is the following statement true, false, or uncertain? "Black Mirror" is popular.
</question>

<options>
(A) True
(B) False
(C) Uncertain
</options>

<nl_cot>
Step 1: Given that \"Stranger Things\" is a popular Netflix show and if a Netflix show is popular, Karen will binge-watch it, we know that Karen would binge-watch \"Stranger Things\" if it were popular. (*@\textcolor{red}{$\Rightarrow$Correct.}@*)

Step 2: However, the popularity of \"Stranger Things\" does not provide any direct information about the popularity of \"Black Mirror.\" (*@\textcolor{red}{$\Rightarrow$Correct.}@*)

Step 3: We know that \"Black Mirror\" is a Netflix show, but we do not have any information that directly links its popularity to Karen's actions. (*@\textcolor{red}{$\Rightarrow$Correct.}@*)

Step 4: The statement that if Karen binge-watches a Netflix show, she will download it, and the fact that Karen does not download \"Black Mirror,\" imply that Karen does not binge-watch \"Black Mirror.\" (*@\textcolor{red}{$\Rightarrow$Correct.}@*)

Step 5: Since Karen does not binge-watch \"Black Mirror,\" and if a show is popular, Karen will binge-watch it, we cannot conclude that \"Black Mirror\" is popular based on Karen's actions. (*@\textcolor{red}{$\Rightarrow$Incorrect. Fail to utilize Transposition.}@*) 

Step 6: The conclusion \"Black Mirror\" is popular cannot be definitively supported or refuted with the given premises. (*@\textcolor{red}{$\Rightarrow$Incorrect.}@*) 

<end_of_nl_cot>
<answer>
The final answer is (C).  (*@\textcolor{red}{$\Rightarrow$Incorrect.}@*) 
<end_of_answer>"

<truth_table>
Let:
    P: \"Stranger Things\" is a popular Netflix show.
    Q: \"Black Mirror\" is a popular Netflix show.
    B: Karen will binge-watch a Netflix show.
    D: Karen will download a Netflix show.
    S: Karen will share a Netflix show with Lisa.
Premises:
    P = True (Given) (*@\textcolor{red}{$\Rightarrow$Correct.}@*)
    If P, then B (Given) (*@\textcolor{red}{$\Rightarrow$Correct.}@*)
    If B, then D (Given) (*@\textcolor{red}{$\Rightarrow$Correct.}@*)
    If and only if B, then D (Given) (*@\textcolor{red}{$\Rightarrow$Correct.}@*)
    \"Black Mirror\" is a Netflix show (Given) (*@\textcolor{red}{$\Rightarrow$Correct.}@*)
    If B, then S (Given) (*@\textcolor{red}{$\Rightarrow$Correct.}@*)
Conclusion to Evaluate:
    Q (Is \"Black Mirror\" popular?) (*@\textcolor{red}{$\Rightarrow$Correct.}@*)
Logical Chain:
    From (5), \"Black Mirror\" is a Netflix show. (*@\textcolor{red}{$\Rightarrow$Correct.}@*)
    From (2) and (3), if \"Black Mirror\" is popular (Q), then Karen will binge-watch it (B). (*@\textcolor{red}{$\Rightarrow$Correct.}@*)
    From (4), if Karen binge-watches \"Black Mirror,\" then she will download it (D). (*@\textcolor{red}{$\Rightarrow$Correct.}@*)
    However, from (6), we know that Karen does not download \"Black Mirror,\" so D = False. (*@\textcolor{red}{$\Rightarrow$Correct.}@*)
    Since B implies D (from (3) and (4)), and D is False, B must also be False. (*@\textcolor{red}{$\Rightarrow$Correct.}@*)
    Since B is False (Karen does not binge-watch \"Black Mirror\"), Q (popularity of \"Black Mirror\") cannot be True. (*@\textcolor{red}{$\Rightarrow$Correct.}@*)
Truth Table (Key Possibilities):
| Q (Q: \"Black Mirror\" is popular) | B (B: Karen will binge-watch \"Black Mirror\") | D (D: Karen will download \"Black Mirror\") |
| False                            | False                                      | False                                    |
| True                             | True                                       | True                                     |
Given that D is False, Q must be False. (*@\textcolor{red}{$\Rightarrow$Correct.}@*)
<end_of_truth_table>
<answer>
The final answer is (B). (*@\textcolor{red}{$\Rightarrow$Correct.}@*)
<end_of_answer>


<code>
from typing import Optional

class Show:
    def __init__(self, name: str, is_popular: Optional[bool] = None, is_downloaded: Optional[bool] = None, is_binged: Optional[bool] = None):
        self.name = name
        self.is_popular = is_popular
        self.is_downloaded = is_downloaded
        self.is_binged = is_binged

def apply_premises(show: Show) -> bool:
    changed = False

    # Premise 1: "Stranger Things" is a popular Netflix show.
    if show.name == "Stranger Things": (*@\textcolor{red}{$\Rightarrow$Correct.}@*) 
        if show.is_popular is None:
            show.is_popular = True
            changed = True

    # Premise 2: If a Netflix show is popular, Karen will binge-watch it.
    if show.is_popular is True and show.name != "Black Mirror": (*@\textcolor{red}{$\Rightarrow$Correct.}@*) 
        if show.is_binged is None:
            show.is_binged = True
            changed = True

    # Premise 3: If and only if Karen binge-watches a Netflix show, she will download it.
    if show.is_binged is True:  (*@\textcolor{red}{$\Rightarrow$Correct.}@*) 
        if show.is_downloaded is None:
            show.is_downloaded = True
            changed = True

    # Premise 4: "Black Mirror" is a Netflix show. (*@\textcolor{red}{$\Rightarrow$Incorrect. 1. Do not encode Transposition. 2. It directly set $\mathrm{show.is\_popular = False}.$, so even if it set it correctly, as it do not include Transposition, code approach still cannot work correctly. }@*) 
    if show.name == "Black Mirror":
        if show.is_popular is None:
            show.is_popular = False
            changed = True

    # Premise 5: If Karen binge-watches a Netflix show, she will share it with Lisa.
    if show.is_binged is True: (*@\textcolor{red}{$\Rightarrow$Correct.}@*) 
        if show.is_downloaded is True:
            if show.name == "Black Mirror":
                if show.is_popular is False:
                    show.is_popular = True
                    changed = True

    return changed

def run_inference(show: Show):
    while apply_premises(show):
        pass

def check_conclusion(show: Show) -> str:
    run_inference(show)
    if show.name == "Black Mirror":
        if show.is_popular is True:
            return "True"
        elif show.is_popular is False:
            return "False"
        else:
            return "Uncertain"
    else:
        return "Uncertain"

def func():
    show = Show(name="Black Mirror", is_popular=None, is_downloaded=None, is_binged=None)
    return check_conclusion(show)

if __name__ == '__main__':
    result = func()
    print("Conclusion: 'Black Mirror' is popular?", result)

<end_of_code>
<answer>
The final answer is (C). (*@\textcolor{red}{$\Rightarrow$Incorrect.}@*) 
<end_of_answer>
\end{promptbox}

\begin{promptbox}[Examples uniquely solved by Truth Table ]

<premises>
All of Peter's pets that are mammals are also animals.  
All monkeys are mammals.  
Peter's pets are all either monkeys or birds.  
Peter's birds can fly.  
All animals that are Peter's pets can breathe.  
If Peter's pet can fly, then it has wings.  
Rock is Peter's pet.  
Rock can fly, or Rock is a bird, or Rock cannot breathe.
</premises>

<conclusion>
Rock is a monkey.
</conclusion>

<question>
Is the following statement true, false, or uncertain? Rock is a monkey.
</question>

<options>
(A) True
(B) False
(C) Uncertain
</options>

<nl_cot>
Step 1: From the premise \"All of Peter's pets that are mammals are also animals,\" we know that any mammal pet of Peter's is an animal. 

Step 2: From the premise \"All monkeys are mammals,\" we know that if something is a monkey, it is also a mammal.

Step 3: From the premise \"Peter's pets are all either monkeys or birds,\" we know that Rock, being a pet of Peter's, must be either a monkey or a bird.

Step 4: From the premise \"Peter's birds can fly,\" we know that if something is a bird, it can fly.

Step 5: From the premise \"All animals that are Peter's pets can breathe,\" we know that any pet of Peter's, whether a bird or a mammal, can breathe.

Step 6: From the premise \"If Peter's pet can fly, then it has wings,\" we know that if a pet of Peter's can fly, it must have wings.

Step 7: From the premise \"Rock can fly, or Rock is a bird, or Rock cannot breathe,\" we have a disjunction. We need to evaluate each part of this disjunction:\n- If Rock can fly, then it must have wings (Step 6), and it must be a bird (Step 4), because only birds can fly among Peter's pets.\n- If Rock is a bird, it can fly (Step 4), and it can breathe (Step 5).\n- If Rock cannot breathe, this contradicts the fact that all of Peter's pets can breathe (Step 5). (*@\textcolor{red}{$\Rightarrow$InCorrect. If Rock can fly, then it must have wings (Step 6), and it must be a bird (Step 4): this step incorrectly utilizes rule fly -> birds to derive rock must be a bird. }@*)

Step 8: Given the disjunction in Step 7, the only consistent option is that Rock is a bird, because if Rock can fly, it must be a bird, and if Rock cannot breathe, it contradicts the premises. Therefore, Rock is a bird.\n\nStep 9: Since Rock is a bird, and all of Peter's pets are either monkeys or birds (Step 3), Rock cannot be a monkey.

<end_of_nl_cot>
<answer>
The final answer is (B).
<end_of_answer>




<truth_table>
Let:  
    M: Rock is a monkey  
    B: Rock is a bird  
    F: Rock can fly  
    A: Rock is an animal  
    W: Rock has wings  
    R: Rock can breathe  

Premises:  
    1. All of Peter's pets that are mammals are also animals. (Not directly about Rock)  
    2. All monkeys are mammals. (Not directly about Rock)  
    3. Peter's pets are all either monkeys or birds. (Not directly about Rock)  
    4. Peter's birds can fly. (B → F)  
    5. All animals that are Peter's pets can breathe. (A → R)  
    6. If Peter's pet can fly, then it has wings. (F → W)  
    7. Rock is Peter's pet.  
    8. Rock can fly, or Rock is a bird, or Rock cannot breathe. (F ∨ B ∨ ¬R)

Conclusion to Evaluate:
    Rock is a monkey (M)

(*@\textcolor{red}{$\Rightarrow$ \textbf{Correct.} The model correctly produce two cases that lead to different conclusions, \emph{i.e.,} True and False. Therefore, it finally produces Uncertain as the final answer. }@*)

Truth Table (Key Possibilities):

| M (Monkey) | B (Bird) | F (Can Fly) | A (Animal) | W (Wings) | R (Can Breathe) |
| True       | False    | True        | True       | True      | True            |
| False      | True     | True        | True       | True      | True            |
<end_of_truth_table>
<answer>
The final answer is (C).
</end_of_answer>

<code>
from typing import Optional

class Pet:
    def __init__(self,
                 is_monkey: Optional[bool] = None,
                 is_bird: Optional[bool] = None,
                 can_fly: Optional[bool] = None,
                 has_wings: Optional[bool] = None,
                 is_animal: Optional[bool] = None):
        self.is_monkey = is_monkey
        self.is_bird = is_bird
        self.can_fly = can_fly
        self.has_wings = has_wings
        self.is_animal = is_animal

def apply_premises(pet: Pet) -> bool:
    changed = False

    # Premise 1: All of Peter's pets that are mammals are also animals.
    # Since all of Peter's pets are either monkeys or birds, and monkeys are mammals,
    # this premise does not directly affect the state of the pet.
    
    # Premise 2: All monkeys are mammals.
    if pet.is_monkey is True and pet.is_animal is not True:
        pet.is_animal = True
        changed = True

    # Premise 3: Peter's pets are all either monkeys or birds.
    if pet.is_monkey is None and pet.is_bird is None:
        pet.is_monkey = False
        pet.is_bird = False
        changed = True

    # Premise 4: Peter's birds can fly.
    if pet.is_bird is True and pet.can_fly is not True:
        pet.can_fly = True
        changed = True

    # Premise 5: All animals that are Peter's pets can breathe.
    if pet.is_animal is True and pet.can_fly is not True:
        pet.can_fly = True
        changed = True

    # Premise 6: If Peter's pet can fly, then it has wings.
    if pet.can_fly is True and pet.has_wings is not True:
        pet.has_wings = True
        changed = True

    # Premise 7: Rock is Peter's pet.
    # This does not change the state of the pet but indicates that the pet is Rock.
    
    # Premise 8: Rock can fly, or Rock is a bird, or Rock cannot breathe.
    if pet.name == "Rock":
        if pet.can_fly is None:
            pet.can_fly = True
            changed = True
        if pet.is_bird is None:
            pet.is_bird = True
            changed = True
        if pet.can_fly is True and pet.has_wings is not True:
            pet.has_wings = True
            changed = True
    (*@\textcolor{red}{$\Rightarrow$ \textbf{Incorrect.} The model struggles to correctly interpret premises that involve multiple possible branches of reasoning, such as disjunctions ($\lor$). It often commits prematurely to one or more branches instead of preserving logical uncertainty.}@*)

    return changed

def run_inference(pet: Pet):
    while apply_premises(pet):
        pass

def check_conclusion(pet: Pet) -> str:
    run_inference(pet)
    if pet.name == "Rock" and pet.is_monkey is True:
        return "True"
    else:
        return "False"

def func():
    pet = Pet(
        is_monkey=None,
        is_bird=None,
        can_fly=None,
        has_wings=None,
        is_animal=None
    )
    pet.name = "Rock"
    return check_conclusion(pet)

if __name__ == '__main__':
    result = func()
    print("Conclusion: Rock is a monkey?", result)
<end_of_code>
<answer>
The final answer is (B).
</end_of_answer>


\end{promptbox}

\end{document}